\documentclass[11pt]{article}

\usepackage[preprint]{acl}

\usepackage{times}
\usepackage{latexsym}

\usepackage[T1]{fontenc}

\usepackage[utf8]{inputenc}

\usepackage{microtype}

\usepackage{inconsolata}

\usepackage{graphicx}
\usepackage{subcaption}
\usepackage{amsmath}
\usepackage{bm}
\usepackage{amssymb}

\usepackage{float}
\usepackage{placeins}

\usepackage{tcolorbox}
\usepackage{fancyvrb,xcolor}
\tcbuselibrary{listingsutf8} %
\usepackage{listings}

\title{Tracing Relational Knowledge Recall in Large Language Models}

\author{Nicholas Popovič \and Michael Färber\\
TU Dresden \& ScaDS.AI Dresden/Leipzig, Germany\\
\small\texttt{\{nicholas.popovic,michael.faerber\}@tu-dresden.de}}

\begin{document}
\maketitle
\begin{abstract}
We study how large language models recall relational knowledge during text generation, with a focus on identifying latent representations suitable for relation classification via linear probes.
Prior work shows how attention heads and MLPs interact to resolve subject, predicate, and object, but it remains unclear which representations support faithful linear relation classification and why some relation types are easier to capture linearly than others.
We systematically evaluate different latent representations derived from attention head and MLP contributions, showing that per-head attention contributions to the residual stream are comparatively strong features for linear relation classification.
Feature attribution analyses of the trained probes, as well as characteristics of the different relation types, reveal clear correlations between probe accuracy and relation specificity, entity connectedness, and how distributed the signal on which the probe relies is across attention heads.
Finally, we show how token-level feature attribution of probe predictions can be used to reveal probe behavior in further detail.
Code is available online\footnote{\url{https://nicpopovic.com/publications/tracing}}.
\end{abstract}

\section{Introduction}
\begin{figure}[!ht]
    \centering
    \includegraphics[width=\linewidth]{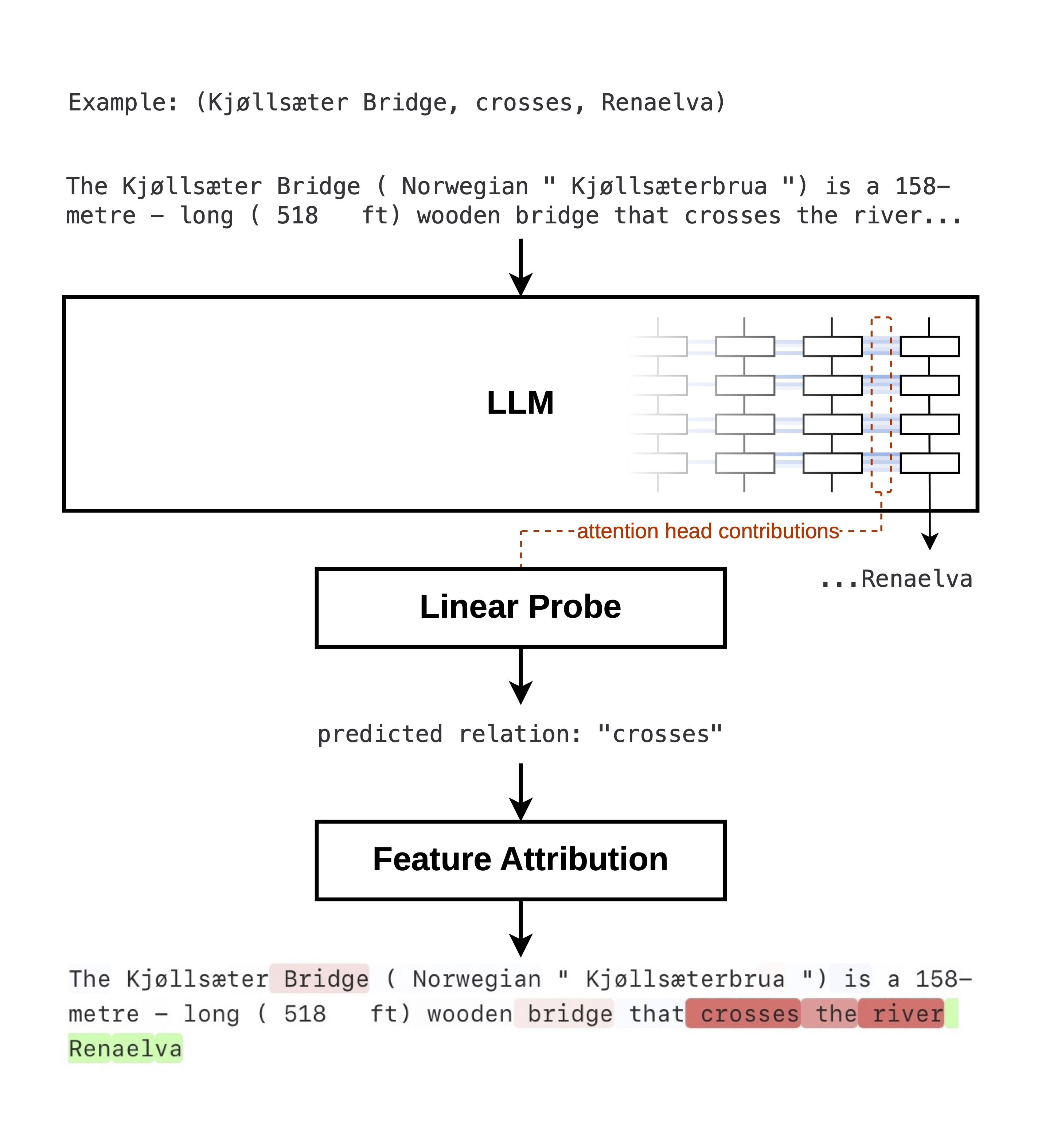}
    \caption{Illustration of attention head contribution probing and token-level feature attribution showing the most influential tokens affecting the probe's prediction.}
    \label{fig:hero}
\end{figure}
Understanding the exact mechanisms by which large language models (LLMs) store and recall relational knowledge is a key area of interpretability research.
Analyzing information flow while generating an appropriate object given a specific subject and predicate has resulted in substantial insight into knowledge representation in LLMs (e.g., \cite{meng2022locating,geva-etal-2023-dissecting,hernandez2024linearity}).

\paragraph{}
State-of-the-art research suggests that when generating an appropriate object for a statement based on a specific subject and predicate: (1) LLMs accumulate information about the \textit{subject entity} in the representations of the middle to late layers at the last token position of the subject span \cite{meng2022locating,geva-etal-2023-dissecting,popovic-farber-2024-embedded,sakata-etal-2025-entity}.
(2) Information about the specific \textit{predicate or relation} in question is accumulated via attention heads into the earlier layers at the token position just prior to the generation of the object \cite{geva-etal-2023-dissecting,liPMETPreciseModel2024,lvInterpretingKeyMechanisms2024,yu-etal-2025-back}.
(3) The appropriate \textit{object entity} is retrieved via attention from MLP sublayers at multiple previous token positions \cite{meng2022locating,geva-etal-2023-dissecting,liPMETPreciseModel2024,yu-ananiadou-2024-neuron} by a process which can sometimes be well-approximated linearly \cite{hernandez2024linearity} and can be traced back to relation-specific neurons \cite{liu-etal-2025-relation-specific}.
Recent work refines this picture, showing that retrieval is distributed across multiple, additively combining mechanisms spanning subject, relation, and last-token positions \cite{chughtai2024summing,yu-etal-2025-back}.
Further research has shown that, for entities, representations are observable via probes with sufficient reliability to enable named entity recognition during generation \cite{popovic-farber-2024-embedded,morandToMMeREfficientEntity2025} and that representations are precise enough for entity disambiguation \cite{sakata-etal-2025-entity}.

\paragraph{}
The aim of this work is to investigate which internal representations support reliable relation classification during generation, and what determines when such probing is faithful.
Unlike entity representations, which are well localized and readily probeable, relation-type features that can be attributed back to individual source tokens have not yet been isolated.
We address this gap by systematically comparing candidate representations derived from attention and MLP sublayers, and by analyzing both the trained probes and the relation types themselves to identify factors that predict probe performance.
Our overarching research questions are: \textit{(i) What are suitable latent representations to probe in an LLM for relation classification?} and \textit{(ii) What properties of a relation and of the trained probe predict when such probing will succeed?} To this end, we make the following contributions: (1) We show that probing per-head attention contribution features is a strong and interpretable basis among latent LLM representations for few-shot relation classification across four instruction-tuned LLMs. (2) We introduce HeadScore and TokenScore, two probe-side attribution methods that decompose each prediction of a trained linear probe over attention heads and over source tokens.
(3) Using these, together with dataset- and relation-level statistics, we identify correlates of linear probe performance.
At the relation level, probe precision is predicted by predicate output range, mean entity connectedness, and example-level lexical similarity.
At the probe level, it is predicted by how concentrated the probe-side signal is across attention heads.
The latter is a property of the trained probe rather than of the underlying model, and may serve as a data-free diagnostic of probe behavior in relation classification.
(4) We use TokenScore to examine whether probe decisions are driven by lexical cues, and find that only a small fraction of errors are consistent with lexical shortcutting, suggesting that linear relation probing is not primarily lexically driven.
We validate these findings across four instruction-tuned LLMs from two model families (LLaMA3 \cite{llama3modelcard} and Qwen3 \cite{qwen3}) ranging from 1 billion to 8 billion parameters.

\section{Related Work}

\subsection{Mechanisms of Knowledge Recall in Large Language Models}
A growing body of work studies how factual and relational knowledge is stored and retrieved in LLMs, and how this process can be localized to specific components that participate in structured recall.
A canonical picture has emerged in which subject information is accumulated at the final token of the subject span \cite{meng2022locating,geva-etal-2023-dissecting}, relation information is propagated to later positions via attention \cite{geva-etal-2023-dissecting,lvInterpretingKeyMechanisms2024}, and object information is retrieved via attention from MLP sublayers \cite{meng2022locating,liPMETPreciseModel2024,yu-ananiadou-2024-neuron,liu-etal-2025-relation-specific}.
Recent work refines this picture, showing that factual recall is driven by several additively combining mechanisms \cite{chughtai2024summing,yu-etal-2025-back}, that relation information can be isolated from subject information at the final token and transplanted across prompts \cite{wang-etal-2024-locating}, and that a small set of attention heads transports the task representation to the recall position \cite{todd2024function,hendel-etal-2023-context}.
For some relation types this process is well approximated by linear transformations, allowing relational information to be decoded from intermediate activations \cite{hernandez2024linearity}.
However, prior work does not identify what distinguishes relations for which such linear approximations are faithful from those for which they fail.
We show that relation specificity, approximated by predicate output range, predicts both probe performance and the concentration of relation-specific features in individual attention heads.

\paragraph{}
Recent work further shows that relation-specific information can be isolated at the level of individual neurons, largely in MLP layers \cite{liu-etal-2025-relation-specific}.
Other analyses highlight the role of attention heads in routing information during recall \cite{geva-etal-2023-dissecting,lvInterpretingKeyMechanisms2024,chughtai2024summing}, and broader surveys suggest that many attention heads specialize in distinct semantic or syntactic roles \cite{attentionheadsurvey2024}.
\citet{wang-etal-2025-tracing} extend this to multi-token reasoning, identifying successive recall stages across heads and MLPs.
However, these results do not directly address whether and how relation-specific signals can be linked back to individual source tokens during generation using a trained probe.
We focus on per-head, per-token attention contributions as probe features.
This makes it possible to relate predictions back to the specific tokens that support them.

\subsection{Information Extraction from Internal Representations}

Recent work has also explored using internal representations of LLMs for information extraction during generation.
For named entity recognition, \citet{popovic-farber-2024-embedded} show that named entities can be extracted during generation using lightweight probes on intermediate states, enabling streaming NER without modifying model parameters.
Follow-up work by \citet{morandToMMeREfficientEntity2025} improves efficiency and robustness.
Relation classification from internal representations remains more challenging: Existing analyses of relational recall \cite{liu-etal-2025-relation-specific,chughtai2024summing} do not provide a probing mechanism that traces relation predictions jointly to attention heads and source tokens during generation.
Our work moves in this direction by combining head-level and token-level attribution on top of a trained linear probe.

\subsection{Attribution for Trained Probes}
In this work we introduce HeadScore and TokenScore, which are closely related to prior work on linear attribution and probing (e.g., \cite{alain2017understanding}).
The resulting decomposition is also conceptually similar to gradient-based attribution methods \cite{pmlr-v70-sundararajan17a} and shares similarities with methods that express predictions as additive contributions from input tokens \cite{james2018beyond}, as well as approaches that analyze contrastive directions in representation space \cite{pmlr-v80-kim18d}.
Methodologically, the closest precedent is \citet{chughtai2024summing}, who extend direct logit attribution to split an attention head's output over individual source tokens in order to study the model's own factual-recall behavior.
TokenScore applies a related idea to a task-specific trained probe, so that attributions explain the probe's decision rather than the model's next-token prediction.

\section{Tracing Relational Knowledge Recall}
\label{sec:tracing}
This section defines the internal features we probe in our experiments.
While prior work identifies attention and MLP sublayers as carriers of relation-specific information, it remains unclear how these signals decompose over source tokens.
We therefore focus on attention heads, which aggregate token information, and define contribution-based features.
Appendix \ref{sec:appendix-probing-illustrations} includes a diagram (Figure \ref{fig:llamablock}) indicating the two locations at which we probe representations.
Finally, in \ref{sec:tokenscore}, we describe our method for identifying the most relevant tokens in the recall of specific relational knowledge.

\subsection{Attention Features}
\label{sec:tracing-attention}

\paragraph{Notation.}
Let $V_h(j)$ denote the value vector produced by attention head $h$ at source position $j$ (obtained by applying the head's value projection to the hidden state at $j$), and let $\mathrm{Attn}_h(t,j)$ denote the softmax attention weight assigned by head $h$ from target position $t$ to source position $j$.
Let $W_{O,h}$ be the slice of the output projection matrix associated with head $h$.

Throughout, $e_1$ and $e_2$ denote the first-mentioned and second-mentioned entity spans in the prompt text.
This means that either can be subject or object, depending on the order in which they are mentioned. $\Delta$ denotes a contribution to the attention residual stream at a fixed target position $t$, which we take to be the relation knowledge recall position (i.e. the position right before the first token of the second entity $e_2$ is generated).

\paragraph{1. Per-head attention contribution ($\Delta_{\text{att},h}$).}
The contribution of attention head $h$ to the residual stream at position $t$ is given by
\[
\Delta_{\text{att},h}(t) = W_{O,h} \left( \sum_{j} \mathrm{Attn}_h(t,j)\, V_h(j) \right).
\]
This corresponds to the contribution of head $h$ to the attention residual stream at position $t$ after aggregating information from all source positions (i.e. previous tokens).

\paragraph{2. Per-head contribution attributable to a specific source token ($\Delta_{\text{att},e_1,h}$).}
We can further decompose the head contribution into terms attributable to individual source positions.
For an arbitrary source token $j$, the contribution of position $j$ via head $h$ to the residual stream at $t$ is
\[
\Delta_{\text{att},h}(t,j) = W_{O,h}\!\left( \mathrm{Attn}_h(t,j)\, V_h(j) \right).
\]

In our experiments, we are primarily interested in contributions originating from the last token of the first-mentioned entity $e_1$ (typically the subject entity).
We therefore define the shorthand
\[
\Delta_{\text{att},e_1,h} \;\triangleq\; \Delta_{\text{att},h}\!\left(t, j_{e_1}\right),
\]
where $j_{e_1}$ denotes the final token of the span corresponding to entity $e_1$.

\paragraph{3. Total contribution from a source token across heads ($\Delta_{\text{att},e_1}$).}
Summing the per-head contributions yields the total contribution of an arbitrary source token $j$ to the attention residual stream at position $t$:
\[
\Delta_{\text{att}}(t,j) \;=\; \sum_{h} \Delta_{\text{att},h}(t,j).
\]

As before, we define the shorthand
\[
\Delta_{\text{att},e_1} \;\triangleq\; \Delta_{\text{att}}\!\left(t, j_{e_1}\right) \;=\; \sum_{h} \Delta_{\text{att},e_1,h}.
\]

\subsection{MLP Features}
\label{sec:tracing-mlp}

MLP features are obtained by propagating attention-side contributions defined in Section~\ref{sec:tracing-attention} into the MLP.
Increased sparsity, as well as the introduction of a non-linearity in the MLP make these features potentially interesting.
In order to correctly account for normalization and non-linearities during the forward propagation, RMSNorm coefficients and gate values are frozen during the regular forward pass and applied to the individual contributions (see Figure \ref{fig:llamablock} in Appendix \ref{sec:appendix-probing-illustrations} for clarification).

In direct analogy to the attention features defined in Section~\ref{sec:tracing-attention}, we define head-specific and entity-specific variants by propagating the corresponding attention contributions.
Specifically, propagating \(\Delta_{\text{att},h}(t)\) yields \(\Delta_{\text{MLP},h}\), while propagating \(\Delta_{\text{att},e_1,h} = \Delta_{\text{att},h}(t,j_{e_1})\) yields \(\Delta_{\text{MLP},e_1,h}\).
Summing over heads gives \(\Delta_{\text{MLP},e_1} = \sum_h \Delta_{\text{MLP},e_1,h}\).

\subsection{Linear Probing \& Feature Attribution}
\label{sec:tokenscore}

We use linear probes so that probe decisions can be decomposed directly over the traced features, enabling both head-level and token-level attribution.
Next, we define a probe-side attribution view at two resolutions.
\emph{HeadScore} is a head-level attribution method that measures how much net signal each attention head contributes to the probe's predicted relation.
\emph{TokenScore} refines the same probe-side signal to the level of source tokens.

\paragraph{Setup.}
We assume a linear probe trained on a subset of per-head attention contributions \(\Delta_{\text{att},h}(t)\), as defined in Section~\ref{sec:tracing-attention}.
Each probe input feature is a single scalar
\[
x_m = \bigl[\Delta_{\text{att},h_m}(t)\bigr]_{d_m},
\]
indexed by \(m \mapsto (\ell_m, h_m, d_m)\), where \(\ell_m\) denotes the layer, \(h_m\) the attention head, and \(d_m\) the model dimension.

Given an input feature vector \(x \in \mathbb{R}^M\), the probe produces class logits
\[
\mathrm{logit}_c = b_c + \sum_{m=1}^M W_{c,m}\, x_m,
\]
with weights \(W \in \mathbb{R}^{C \times M}\) and bias \(b \in \mathbb{R}^C\).

\paragraph{Contrast direction.}
For a given example, let \(\hat{c}\) denote the predicted class.
We define a contrast direction by comparing \(\hat{c}\) against a softmax-weighted mixture of all remaining classes:
\[
\Delta W \;=\; W_{\hat{c}} \;-\; \sum_{c \neq \hat{c}} \pi_c\, W_c,
\]
\[
\qquad \pi_c = \frac{\exp(\mathrm{logit}_c / \tau)} {\sum_{c' \neq \hat{c}} \exp(\mathrm{logit}_{c'} / \tau)}.
\]
This yields a vector \(\Delta W \in \mathbb{R}^M\) that captures how each probe feature contributes to separating the predicted class from all competing classes.
We use \(\tau = 1\) throughout.

\paragraph{Head attribution.}
The direct signed contribution of probe feature \(m\) to this prediction contrast is \(\Delta W_m x_m\).
Aggregating these feature-level contributions by attention head yields the HeadScore
\[
\mathrm{HeadScore}_{\ell,h} = \sum_{m:\,\ell_m=\ell,\,h_m=h} \Delta W_m \, x_m.
\]

\paragraph{Token attribution.}
HeadScore resolves attribution only to heads.
To localize the same probe-side signal to source tokens, we use the token decomposition of each head contribution (Section~\ref{sec:tracing}):
\[
\Delta_{\text{att},h}(t) = \sum_j \Delta_{\text{att},h}(t,j).
\]
Accordingly,
\[
x_m = \bigl[\Delta_{\text{att},h_m}(t)\bigr]_{d_m} = \sum_j \bigl[\Delta_{\text{att},h_m}(t,j)\bigr]_{d_m}.
\]

For a source token \(j\), the TokenScore at layer \(\ell\) is defined as
\[
\mathrm{TokenScore}_\ell(j) = \sum_{m:\,\ell_m=\ell} \Delta W_m \cdot \bigl[\Delta_{\text{att},h_m}(t,j)\bigr]_{d_m}.
\]
We optionally aggregate across layers,
\[
\mathrm{TokenScore}(j) = \sum_{\ell} \mathrm{TokenScore}_\ell(j),
\]
or visualize TokenScores on a per-layer basis.

\paragraph{Interpretation.}
Positive HeadScores or TokenScores indicate evidence in favor of the predicted relation under the linear probe, while negative values indicate opposing evidence.
It is important to note that these attributions explain the probe's decision, not the internal computation performed by the underlying LLM.

\section{Experiments}

\subsection{Simple Few-Shot Relation Classification}
\label{sec:fillinblanks-experiment}
As introduced in Sections \ref{sec:tracing-attention} and \ref{sec:tracing-mlp}, different internal representations and contributions, traceable to individual attention heads, can be accessed along the knowledge recall circuit.
In this set of experiments we explore which of these are best suited for classification between different relation types.
As baselines we also include the full state at the two probing locations, Attention and MLP (without isolated contributions for individual attention heads or tokens).

\paragraph{Experiment Setup}
We examine an $n$-way $k$-shot few-shot relation classification setting to determine whether representations are sufficiently specific to distinguish between different relation types.
Following \citet{liu-etal-2025-relation-specific} (see Appendix \ref{sec:appendix-RelSpec-fewshot}), we use the support set of each episode to calculate expertise scores \cite{pmlr-v162-cuadros22a} and select only the top $m=3000$ features per relation type as a feature vector.
We then train a linear probe for $200$ epochs using cross entropy loss and Adam \cite{kingma2015adam} on the support set and evaluate against a query set.

\begin{figure}[!ht]
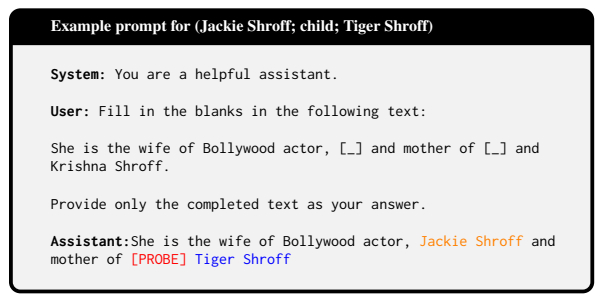

\centering
\begin{tcolorbox}[title=Example prompt for (Jackie Shroff; child; Tiger Shroff), fonttitle=\bfseries\tiny, colback=gray!10, colframe=black, width=1\linewidth, listing only, listing options={
    basicstyle=\ttfamily\small,
    breaklines=true,
    showstringspaces=false
}]
\begin{Verbatim}[commandchars=\\\(\),fontsize=\tiny]
\textbf(System:) You are a helpful assistant.

\textbf(User:) Fill in the blanks in the following text:

She is the wife of Bollywood actor, [_] and mother of [_] and
Krishna Shroff.

Provide only the completed text as your answer.

\textbf(Assistant:)She is the wife of Bollywood actor,\color(orange) Jackie Shroff\color(black) and
mother of \color(red)[PROBE]\color(blue) Tiger Shroff
\end{Verbatim}
\end{tcolorbox}
\caption{\textit{Fill-in-the-blanks} prompt used for evaluation. In red, we show the token position at which we probe for representations. The subject entity is marked in orange, while the object entity is shown in blue. It is not visible to the model during probing.}
\label{fig:llm-prompt-eval}
\end{figure}

\paragraph{Prompt Format}
Prior work studies relational knowledge recall using fixed phrase templates and often a question-answering setting to ensure that the LLM recalls the desired type of knowledge.
Instead, we do not use fixed templates, so it is possible that key information relevant to the predicate occurs after the object entity is mentioned.
As a result of autoregressivity, the LLM then has incomplete information about the predicate when generating the object.
In order to avoid this, we adopt a \textit{fill-in-the-blanks} prompt, shown in Figure \ref{fig:llm-prompt-eval}.
By first providing the final text with masked subject and object spans, all necessary context outside of the object entity itself is available, even in the autoregressive setting\footnote{Additional examples highlighting such cases are given in Figure \ref{fig:appendix-fill-in-blanks-examples} in Appendix \ref{section:appendix-fill-in-blanks-examples}}.

\paragraph{Evaluation}
We use the validation split of FewRel \cite{han-etal-2018-fewrel} for our experiments.
It contains 700 examples each for 16 different relation types and annotated subject and object locations.
Since it is generally considered an easy dataset for few-shot relation classification (usually evaluated in $n \in \{1,5\}, k \in \{1,5\}$), we expect strong performance in this regime; substantially lower accuracies would suggest that our features are not suitable for relation classification.
We filter both the support and query data to only include examples where $e_1$ is the subject and $e_2$ is the object.
All reported results are accuracies across $5$ different random seeds evaluated on $500$ episodes each.
We perform these experiments on 4 different instruction-tuned LLMs from 2 different model families, namely LLaMA-3.2$_\textnormal{\small{1B/3B-Instruct}}$, LLaMA-3.1$_\textnormal{\small{8B-Instruct}}$ \cite{llama3modelcard}, Qwen3$_\textnormal{\small{4B-Instruct}}$ \cite{qwen3}.

\begin{table}[]
    \centering
    \small
    \begin{tabular}{l|cccc}
         & \multicolumn{3}{c|}{LLaMA-3.1/3.2} & Qwen-3  \\
         & 1B-Inst. & 3B-Inst. & 8B-Inst. & 4B-Inst.  \\
        \hline
        MLP & $85.79\%$ & $86.40\%$ & $85.90\%$ & $80.37\%$ \\
        $\Delta_{\text{MLP},h}$  & $89.96\%$ & $90.90\%$ & $89.99\%$ & $88.43\%$ \\
        $\Delta_{\text{MLP},e_1}$ & $66.66\%$ & $66.02\%$ & $65.85\%$ & $68.49\%$ \\
        $\Delta_{\text{MLP},e_1,h}$ & $72.56\%$ & $72.33\%$ & $71.59\%$ & $73.98\%$ \\        
        \hline
        Attention & $83.65\%$ & $86.61\%$ & $86.79\%$ & $75.06\%$ \\
        $\Delta_{\text{att},h}$ & \bm{$90.26\%$} & \bm{$91.06\%$} & \bm{$91.09\%$} & \bm{$89.66\%$} \\
        $\Delta_{\text{att},e_1}$ & $59.64\%$ & $60.16\%$ & $59.85\%$ & $59.58\%$ \\
        $\Delta_{\text{att},e_1,h}$ & $64.96\%$ & $65.21\%$ & $64.59\%$ & $65.29\%$ \\
        \hline
    \end{tabular}
    \caption{Few-shot relation classification accuracy on FewRel validation split for $n=5,k=5$ with constraint $e_1=s$ and $e_2=o$.}
    \label{tab:5-way-5-shot-m-3000-nostd}
\end{table}

\paragraph{Results}
Table \ref{tab:5-way-5-shot-m-3000-nostd} contains the results for the 5-way-5-shot setting.
Further results for more configurations are included in Appendix~\ref{sec:appendix-fewshot}.
Overall, accuracies are high, reaching up to $91.09\%$.
While this range of scores has been achieved on FewRel \cite{baldini-soares-etal-2019-matching}, this shows that the identified features are suitable.
Our results show clearly that both $\Delta_{\text{att},h}$ and $\Delta_{\text{MLP},h}$ are the strongest representations for relation classification, resulting in higher accuracies than the non-attributable baselines Attention and MLP.
On the other hand, training probes on $\Delta_{\text{MLP},e_1}$, $\Delta_{\text{MLP},e_1,h}$, $\Delta_{\text{att},e_1}$, $\Delta_{\text{att},e_1,h}$, results in substantially poorer performance.
In Appendix~\ref{sec:appendix-num-RelSpec} we include results for different amounts of RelSpec features.

\paragraph{Findings}
Accuracies exceeding 90\% can be achieved by probing individual attention head contributions.
This demonstrates that classification is possible, but is below existing methods' accuracies which can exceed 97\% even in the non-simplified version of this task \cite{baldini-soares-etal-2019-matching}.
In Appendix \ref{sec:appendix-fewshot}, we provide further results for different $n$-way-$k$-shot settings.
Among the examined representations, $\Delta_{\text{att},h}$ scores highest for relation classification.
This enables us to trace predictions back to specific attention heads, and by extension to individual tokens, opening the possibility for detailed error analysis.
Propagating contributions to MLP layers did not result in any improved representational abilities.
Observing source token contributions ($\Delta_{\text{att},e_1}, \Delta_{\text{att},e_1, h}$) is not beneficial to relation classification.
As a result of these findings, all following experiments focus on $\Delta_{\text{att},h}$.

\subsection{What Affects Classification Accuracy?}

\begin{figure*}[!ht]
    \centering
    \includegraphics[width=1\linewidth]{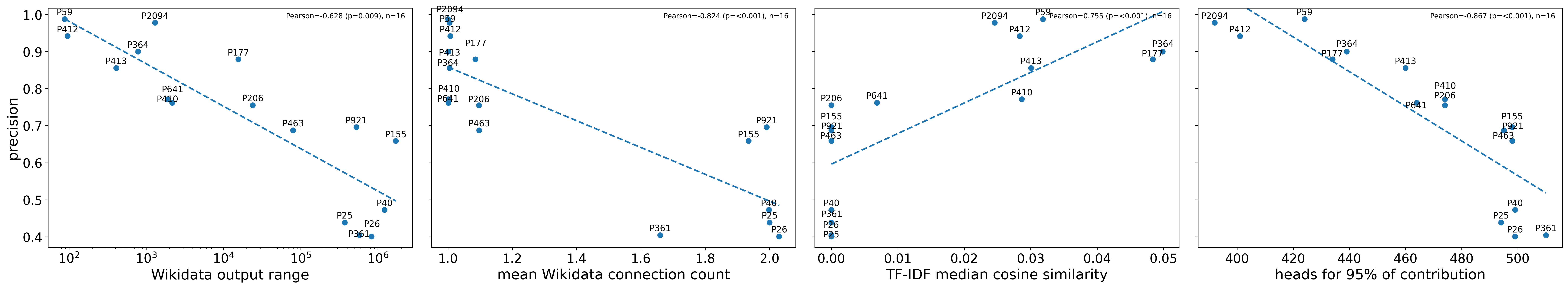}
    \caption{Correlations observed for probe precision (here for LLaMA-3.1$_{\textnormal{\small{8B}}}$). Output range is the number of distinct Wikidata objects for a property; mean Wikidata connection count is the mean number of direct truthy properties connecting $1000$ sampled distinct subject--object pairs for which that property holds. Heads for 95\% of contribution is the minimum number of attention heads whose cumulative absolute contribution reaches 95\% of the total headscores.}
    \label{fig:correlations}
\end{figure*}

After establishing $\Delta_{\text{att},h}$ as the most suitable feature, we increase the task difficulty and perform feature attribution analyses to examine whether we can establish factors influencing probe performance.

\begin{figure}[!ht]
\centering
\begin{tcolorbox}[title=Example prompt for (Liberty Belle; child; Jesse Quick), fonttitle=\bfseries\tiny, colback=gray!10, colframe=black, width=1\linewidth, listing only, listing options={
    basicstyle=\ttfamily\small,
    breaklines=true,
    showstringspaces=false
}]
\begin{Verbatim}[commandchars=\\\(\),fontsize=\tiny]
\textbf(System:) You are a helpful assistant.

\textbf(User:) Fill in the blanks in the following text:

Lance later appears in The Titans as a friend of [_]'s mother,
the aged heroine\color(orange) Liberty Belle\color(black).

Provide only the completed text as your answer.

\textbf(Assistant:) Lance later appears in The Titans as a friend of\color(red)[PROBE]
\color(blue)Jesse Quick
\end{Verbatim}
\end{tcolorbox}
\caption{Example of a prompt with inverted subject-object order and our modified prompt for this situation. Since the object entity (Jesse Quick) now appears before the subject entity, the model would retrieve the subject instead of the object if we were to use two blanks as in the prompt used so far. By using only a single blank for the object, the model again recalls the object while having the full text as context.}
\label{fig:llm-prompt-eval-th}
\end{figure}

\paragraph{Experiment Setup}
We modify the experiment setup by no longer filtering the dataset based on entity order.
Due to the LLM's autoregressivity, this results in a need for further considerations: If the subject and object of a statement are reversed, knowledge recall would now occur for the subject instead of the object, resulting in the inverse relation type applying.
We fix this by modifying the prompt as shown in Figure \ref{fig:llm-prompt-eval-th} for such cases.
Further, we increase $n$ to $16$, resulting in a $16$-way-$5$-shot task.
We report precision, recall, F1 scores.

\begin{table}[h]
\centering
\small
\begin{tabular}{lccc}
\hline
\textbf{Model} & \textbf{Precision (\%)} & \textbf{Recall (\%)} & \textbf{F1 (\%)} \\
\hline
LLaMA-3.2$_{\textnormal{\small{1B}}}$ & 68.02 & 68.17 & 68.03 \\
LLaMA-3.2$_{\textnormal{\small{3B}}}$ & 70.56 & 69.75 & 70.01 \\
LLaMA-3.1$_{\textnormal{\small{8B}}}$ & 72.45 & 72.08 & 72.13 \\
\hline
Qwen3$_{\textnormal{\small{4B}}}$ & 66.84 & 68.10 & 66.81 \\
\hline
\end{tabular}
\caption{Macro-averaged precision, recall, and F1-score (percentages) across all properties for each model in the $16$-way-$5$-shot setting.}
\label{tab:16way-results}
\end{table}

\paragraph{Results}
In Table \ref{tab:16way-results}, we show global scores per LLM, while in Appendix \ref{sec:appendix-16way}, we provide the full per class breakdowns.
The increased task difficulty is reflected clearly in lower F1 scores, ranging from $66.81\%$ to $72.13\%$.
Looking at the full classification report, we further find that per class F1 scores vary widely across relation types (but are similar between models).
For example, for LLaMA-3.1$_{\textnormal{\small{8B}}}$, for "P361 - part of" classification F1 is $39.76\%$, while it is $99.24\%$ for "P59 - constellation".
Below, we present clear correlations which we observe that could be used to predict probe performance for new relation types.

\begin{figure*}[!ht]
    \centering
    \includegraphics[width=1\linewidth]{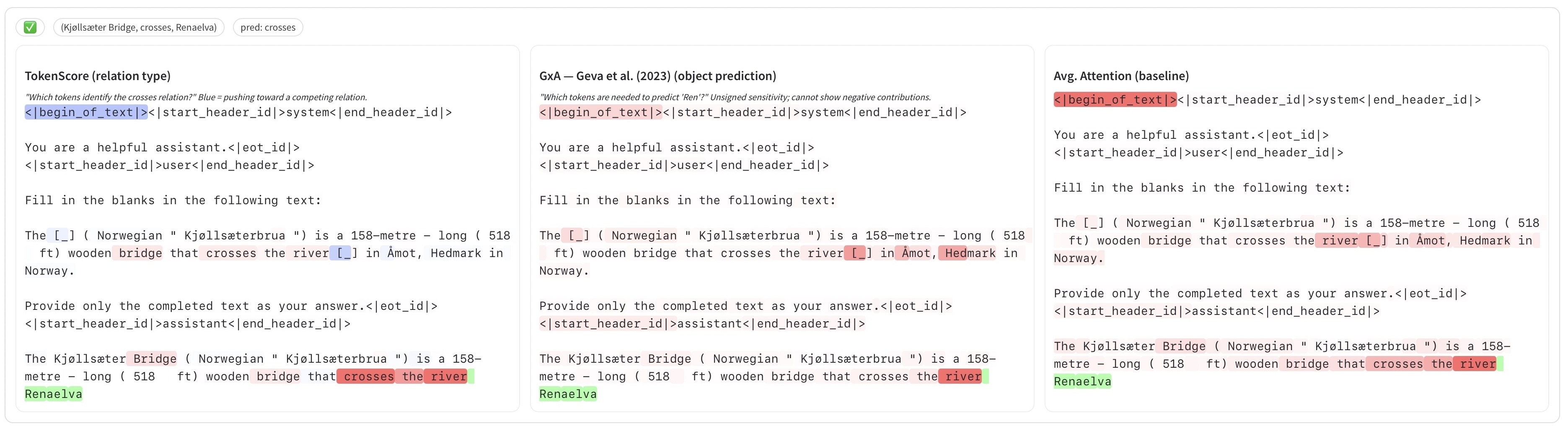}
    \caption{Example of a correct probe prediction for the relation type "crosses" in a $16$-way-$5$-shot setting using LLaMA-3.1$_\textnormal{\small{1B-Instruct}}$. Shown are token-level (coloring normalized per query) TokenScore, GxA (gradient $\times$ activation, following \citet{geva-etal-2023-dissecting}, Appendix B), and avg. attention weights. We observe that TokenScores are high for tokens relevant to the relation type "crosses", such as "crosses the river" or "Bridge".}
    \label{fig:tokenscore_main}
\end{figure*}

\begin{figure*}[!ht]
    \centering
    \includegraphics[width=1\linewidth]{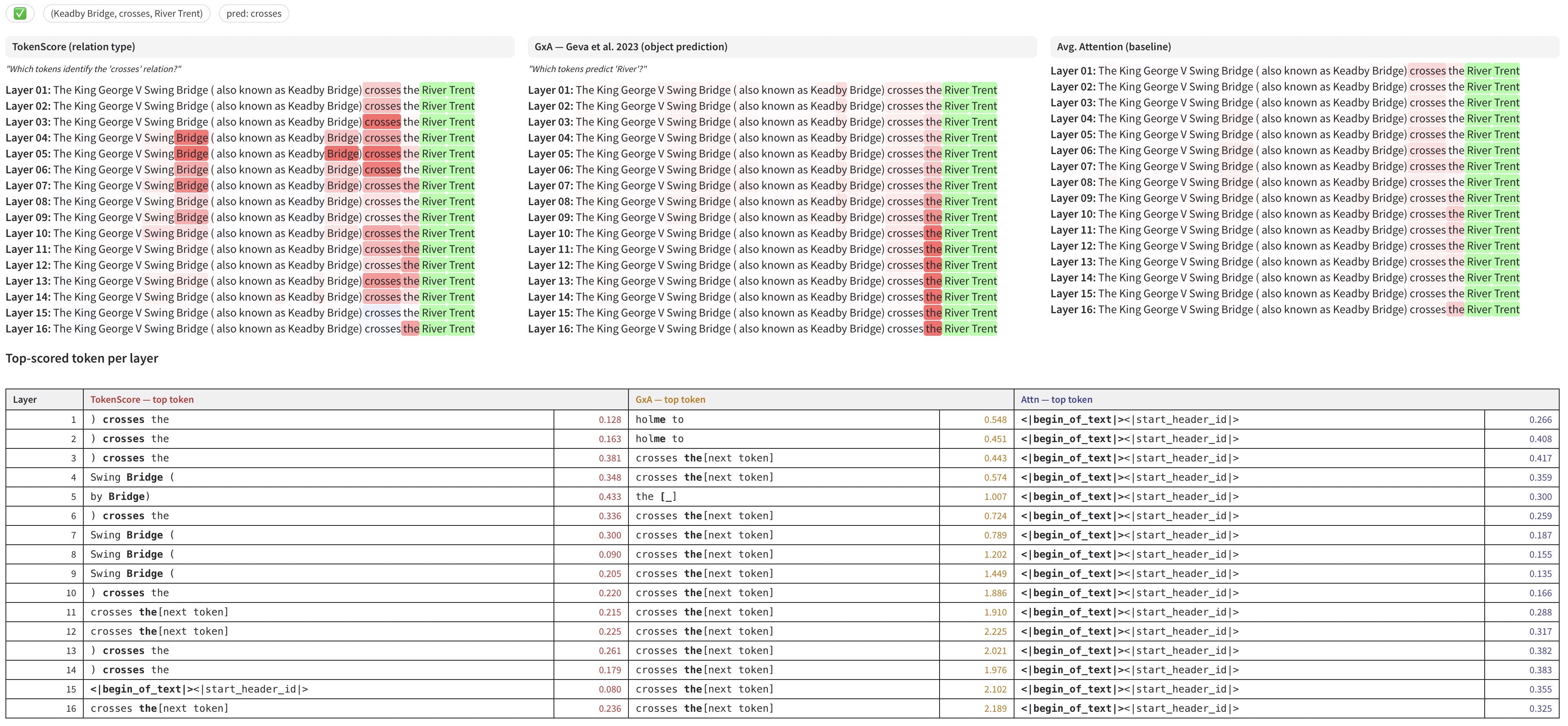}
    \caption{Example of a correct probe prediction for the relation type "crosses" in a $16$-way-$5$-shot setting using LLaMA-3.1$_\textnormal{\small{1B-Instruct}}$. Shown are layerwise (coloring normalized per layer) TokenScore, GxA \cite{geva-etal-2023-dissecting}, and avg. attention weights. Further, we show the top scoring token per layer (and the surrounding context) for each method. We observe that top TokenScores are typically assigned to tokens relevant to the relation type "crosses", such as "crosses" or "bridge".}
    \label{fig:token_attribution_layerwise}
\end{figure*}

\paragraph{Correlation Analysis}
Figure \ref{fig:correlations} shows four correlations.
First, probe precision is lower for relation types with higher Wikidata output range, where output range is the number of distinct objects occurring for a property in Wikidata.
Second, for each relation type, we sample $1000$ distinct subject--object pairs connected by that property and count how many direct truthy Wikidata properties connect the two entities.
Higher mean connection count is associated with lower probe precision.
Third, as a sanity check for a possible lexical confound, we measure the median pairwise cosine similarity between TF-IDF representations of the raw FewRel examples for each relation type.
Relations whose examples are more lexically similar tend to yield higher probe precision.
These first three correlations concern properties of the relation and its textual realizations, and thus characterize the difficulty of the input side.
We then turn to the trained probe itself: ranking attention heads by $\lvert \mathrm{HeadScore}_{\ell,h} \rvert$, we report the minimum number of heads whose cumulative absolute contribution reaches $95\%$ of $\sum_{\ell,h}\lvert \mathrm{HeadScore}_{\ell,h} \rvert$.
Relation types for which this mass is concentrated in fewer heads show higher probe precision than those for which the signal is spread across more heads.

\paragraph{Discussion}
Taken together, these observations suggest several plausible explanations, which our current analysis cannot disentangle.
Higher output range may reflect more generic relations whose mentions are harder for the LLM itself to identify reliably in context, thus harming probe performance.
Likewise, higher entity connectedness may indicate that several relational associations between the same subject-object pair coexist in the LLM's internal representation at the recall position.
If multiple such relations are encoded in overlapping directions in the residual stream, the linear separability on which the probe depends is degraded, even if the model itself can still select the correct object downstream.
This connects to the notion of feature superposition \cite{elhage2022superposition}.
The TF-IDF correlation further raises the possibility that part of the observed effect is driven by recurring lexical cues in the examples, which may be exploited by the probe and/or the LLM; we investigate this possibility directly in Section \ref{sec:tokenlevelattribution}.
Our evidence is otherwise correlational, so distinguishing among the remaining explanations will require controlled interventions and more tightly designed experiments.
Lastly, the probe-side concentration result does not describe the data, but the behavior of the trained probes.
This suggests that HeadScore concentration may be useful as a data-free diagnostic of probe behavior, in settings where labeled evaluation data are limited and in related probing tasks beyond relation classification.
It may also provide a useful signal for grouping relations in more exploratory settings.
We leave these possibilities to future work.

\subsection{Inspecting Token-Level Attribution}
\label{sec:tokenlevelattribution}

Next, we turn to token-level feature attribution, which reveals from which tokens decisive signals informing a probe's prediction originate.
In Figure \ref{fig:tokenscore_main} we show an example of TokenScore and two baselines, gradient-based feature attribution \cite{geva-etal-2023-dissecting} and the average attention weights assigned to each token.
In Figure \ref{fig:token_attribution_layerwise}, we additionally show examples of these scores on a per layer basis, together with the highest scoring token in each layer.
During manual inspection, we observe that TokenScore, more so than the two baselines, is often highest for relation-relevant tokens when the probe's prediction is correct (for example, "crosses" in the figures).
However, we also see elevated TokenScores in tokens that are likely to co-occur with a given relation but not directly describe it (for example, "bridge" in the shown figures), which would indicate a stronger reliance on lexical cues rather than semantics.
Combined with the observed correlations between average probe precision and TF-IDF similarity, we therefore set out to investigate how much trained probes are influenced by lexical cues.

\paragraph{Experiment Setup}
We run the same $16$-way-$5$-shot setting as in the previous section.
For each episode, we construct TF-IDF statistics from the support prompts only.
This yields, for each relation type in the episode, a token-level score indicating how characteristic a token is of that relation relative to the other relations in the same episode.
For each query, we then compare TokenScore to these token-level lexical scores for the probe's predicted relation.

\paragraph{Evaluation}
We report three quantities.
First, we measure the query-level Spearman correlation $\rho$ between per-token TokenScore and the per-token lexical scores induced by the predicted relation.
This captures whether tokens receiving high attribution also tend to be the tokens that are most lexically characteristic of that relation under the support set.
Second, we measure Mass, the fraction of positive TokenScore mass that falls on tokens whose lexical score is positive for the predicted relation.
This captures how much of the probe's positive evidence is concentrated on lexically supportive tokens.
Finally, we measure StrongAlign$_{\times}$, the fraction of incorrect predictions whose TokenScore is strongly aligned with the predicted relation's lexical profile.
We consider TokenScore strongly aligned when Spearman correlation is at least $0.30$ and at least $50\%$ of positive TokenScore mass falls on tokens with positive lexical score for the predicted relation.
This is intended as a conservative indicator of errors that are consistent with lexical shortcutting.

\begin{table}[t]
\centering
\small
\begin{tabular}{lccc}
\hline
\textbf{Model} & \textbf{$\rho$} & \textbf{Mass} & \textbf{StrongAlign$_{\times}$ (\%)} \\
\hline
LLaMA-3.2$_{\textnormal{\small{1B}}}$ & 0.115 & 0.491 & 7.7 \\
LLaMA-3.2$_{\textnormal{\small{3B}}}$ & 0.105 & 0.491 & 5.8 \\
LLaMA-3.1$_{\textnormal{\small{8B}}}$ & 0.095 & 0.475 & 5.3 \\
Qwen3$_{\textnormal{\small{4B}}}$ & 0.099 & 0.490 & 5.9 \\
\hline
\end{tabular}
\caption{Lexical-cue analysis. $\rho$ is the mean query-level Spearman correlation between per-token TokenScore and the predicted relation's episode-local lexical profile. Mass is the mean fraction of positive TokenScore mass that falls on tokens whose lexical score is positive for the predicted relation. StrongAlign$_{\times}$ is the fraction of incorrect predictions whose TokenScore is strongly aligned with the predicted relation's lexical profile.}
\label{tab:tokenscore-lexical-compact}
\end{table}

\paragraph{Results}
The results are shown in Table \ref{tab:tokenscore-lexical-compact}.
Across models, the resulting TokenScore--lexical alignment is low, with mean query-level Spearman correlations between $0.095$ and $0.115$.
Thus, the tokens receiving high probe attribution are only weakly aligned with the tokens that are most lexically characteristic of the predicted relation.
Mass ranges from $0.475$ to $0.491$, indicating that a substantial share of positive TokenScore mass falls on lexically supportive tokens despite the low rank correlation.
Looking at StrongAlign$_{\times}$, we see that only between $5.3\%$ and $7.7\%$ of errors show strong alignment between TokenScore and the lexical profile, suggesting that only a minority of errors can be readily attributed to lexical shortcutting.

\paragraph{Discussion}
Taken together, these results indicate that lexical cues may contribute, but they do not suggest that probe decisions are primarily driven by such cues.
The relation-level correlations with output range and entity connectedness therefore remain plausible potential explanations and are not reduced to a purely lexical explanation.

\section{Conclusion}

We studied which internal representations in LLMs are most suitable for relation classification using linear probes.
Across four instruction-tuned models, per-head attention contribution features $\Delta_{\text{att},h}$ were consistently the strongest probe inputs, outperforming probes on full attention and MLP states.
To analyze these probes, we introduced HeadScore and TokenScore, two probe-side attribution methods that decompose a probe's prediction over attention heads and source tokens.

Increasing task difficulty reveals wide variation in probe performance across relation types, which is correlated negatively with properties of the relation such as output range and entity connectedness as approximated using Wikidata, as well as with the degree of probe-side signal distribution across attention heads.
The latter is a property of the trained probe itself and may serve as a data-free diagnostic of probe behavior beyond the specific relation-classification setting studied here.

We also observe correlations between probe attribution and lexical cues, but token-level analyses suggest that these cues are not the primary driver of probe decisions.
Overall, our results suggest that the success of linear relation classification of LLM representations is heavily relation dependent, and that this dependence is not solely attributable to lexical cues and may also be influenced by output range or by superposition of multiple relational features in the residual stream, a hypothesis we leave for future work.
Our work provides directions and tools for further investigation in this area.

\section*{Limitations}

This work focuses on relation \emph{classification} rather than full relation extraction.
While we analyze representations at a fixed recall position during generation, we do not address the problem of detecting whether a relation is expressed at all, nor do we consider NOTA or negative cases.
Extending the approach to open-world extraction settings remains future work.

Our experiments are conducted on English-language data and a limited set of relations derived from FewRel.
Although we intentionally include both highly specific and generic relations, the results may not generalize to relations with substantially different linguistic realizations or to domains with noisier supervision.
In particular, our use of Wikidata property output range as a proxy for relation specificity is coarse and may not capture all factors that influence representational concentration.

We evaluate a small number of instruction-tuned models from two model families.
While trends are consistent across model sizes and architectures, we do not study very large models or models with substantially different training regimes.
Finally, our analyses rely on linear probes and linear attributions; while these are effective for identifying and tracing relation recall under certain conditions, they may fail to capture more distributed or non-linear mechanisms.

\bibliography{cited-references}

\begin{thebibliography}{28}
\providecommand{\natexlab}[1]{#1}

\bibitem[{Alain and Bengio(2017)}]{alain2017understanding}
Guillaume Alain and Yoshua Bengio. 2017.
\newblock \href {https://openreview.net/forum?id=HJ4-rAVtl} {Understanding intermediate layers using linear classifier probes}.
\newblock In \emph{Proceedings of the 5th International Conference on Learning Representations (ICLR 2017), Workshop Track}, Toulon, France.

\bibitem[{Baldini~Soares et~al.(2019)Baldini~Soares, FitzGerald, Ling, and Kwiatkowski}]{baldini-soares-etal-2019-matching}
Livio Baldini~Soares, Nicholas FitzGerald, Jeffrey Ling, and Tom Kwiatkowski. 2019.
\newblock \href {https://doi.org/10.18653/v1/P19-1279} {Matching the blanks: Distributional similarity for relation learning}.
\newblock In \emph{Proceedings of the 57th Annual Meeting of the Association for Computational Linguistics}, pages 2895--2905, Florence, Italy. Association for Computational Linguistics.

\bibitem[{Chughtai et~al.(2024)Chughtai, Cooney, and Nanda}]{chughtai2024summing}
Bilal Chughtai, Alan Cooney, and Neel Nanda. 2024.
\newblock \href {https://arxiv.org/abs/2402.07321} {Summing up the facts: Additive mechanisms behind factual recall in llms}.
\newblock \emph{Preprint}, arXiv:2402.07321.

\bibitem[{Cuadros et~al.(2022)Cuadros, Zappella, and Apostoloff}]{pmlr-v162-cuadros22a}
Xavier~Suau Cuadros, Luca Zappella, and Nicholas Apostoloff. 2022.
\newblock \href {https://proceedings.mlr.press/v162/cuadros22a.html} {Self-conditioning pre-trained language models}.
\newblock In \emph{Proceedings of the 39th International Conference on Machine Learning}, volume 162 of \emph{Proceedings of Machine Learning Research}, pages 4455--4473. PMLR.

\bibitem[{Dubey et~al.(2024)Dubey, Jauhri, Pandey, Kadian, Al-Dahle, Letman, Mathur, Schelten, Yang, and et~al.}]{llama3modelcard}
Abhimanyu Dubey, Abhinav Jauhri, Abhinav Pandey, Abhishek Kadian, Ahmad Al-Dahle, Aiesha Letman, Akhil Mathur, Alan Schelten, Amy Yang, and Angela~Fan et~al. 2024.
\newblock \href {https://arxiv.org/abs/2407.21783} {The llama 3 herd of models}.
\newblock \emph{Preprint}, arXiv:2407.21783.

\bibitem[{Elhage et~al.(2022)Elhage, Hume, Olsson, Schiefer, Henighan, Kravec, Hatfield-Dodds, Lasenby, Drain, Chen, Grosse, McCandlish, Kaplan, Amodei, Wattenberg, and Olah}]{elhage2022superposition}
Nelson Elhage, Tristan Hume, Catherine Olsson, Nicholas Schiefer, Tom Henighan, Shauna Kravec, Zac Hatfield-Dodds, Robert Lasenby, Dawn Drain, Carol Chen, Roger Grosse, Sam McCandlish, Jared Kaplan, Dario Amodei, Martin Wattenberg, and Christopher Olah. 2022.
\newblock \href {https://arxiv.org/abs/2209.10652} {Toy models of superposition}.
\newblock \emph{Preprint}, arXiv:2209.10652.

\bibitem[{Geva et~al.(2023)Geva, Bastings, Filippova, and Globerson}]{geva-etal-2023-dissecting}
Mor Geva, Jasmijn Bastings, Katja Filippova, and Amir Globerson. 2023.
\newblock \href {https://doi.org/10.18653/v1/2023.emnlp-main.751} {Dissecting recall of factual associations in auto-regressive language models}.
\newblock In \emph{Proceedings of the 2023 Conference on Empirical Methods in Natural Language Processing}, pages 12216--12235, Singapore. Association for Computational Linguistics.

\bibitem[{Han et~al.(2018)Han, Zhu, Yu, Wang, Yao, Liu, and Sun}]{han-etal-2018-fewrel}
Xu~Han, Hao Zhu, Pengfei Yu, Ziyun Wang, Yuan Yao, Zhiyuan Liu, and Maosong Sun. 2018.
\newblock \href {https://doi.org/10.18653/v1/D18-1514} {{F}ew{R}el: A large-scale supervised few-shot relation classification dataset with state-of-the-art evaluation}.
\newblock In \emph{Proceedings of the 2018 Conference on Empirical Methods in Natural Language Processing}, pages 4803--4809, Brussels, Belgium. Association for Computational Linguistics.

\bibitem[{Hendel et~al.(2023)Hendel, Geva, and Globerson}]{hendel-etal-2023-context}
Roee Hendel, Mor Geva, and Amir Globerson. 2023.
\newblock \href {https://doi.org/10.18653/v1/2023.findings-emnlp.624} {In-context learning creates task vectors}.
\newblock In \emph{Findings of the Association for Computational Linguistics: EMNLP 2023}, pages 9318--9333, Singapore. Association for Computational Linguistics.

\bibitem[{Hernandez et~al.(2024)Hernandez, Sharma, Haklay, Meng, Wattenberg, Andreas, Belinkov, and Bau}]{hernandez2024linearity}
Evan Hernandez, Arnab~Sen Sharma, Tal Haklay, Kevin Meng, Martin Wattenberg, Jacob Andreas, Yonatan Belinkov, and David Bau. 2024.
\newblock \href {https://openreview.net/forum?id=w7LU2s14kE} {Linearity of relation decoding in transformer language models}.
\newblock In \emph{The twelfth international conference on learning representations}.

\bibitem[{Kim et~al.(2018)Kim, Wattenberg, Gilmer, Cai, Wexler, Viegas, and sayres}]{pmlr-v80-kim18d}
Been Kim, Martin Wattenberg, Justin Gilmer, Carrie Cai, James Wexler, Fernanda Viegas, and Rory sayres. 2018.
\newblock \href {https://proceedings.mlr.press/v80/kim18d.html} {Interpretability beyond feature attribution: Quantitative testing with concept activation vectors ({TCAV})}.
\newblock In \emph{Proceedings of the 35th International Conference on Machine Learning}, volume~80 of \emph{Proceedings of Machine Learning Research}, pages 2668--2677. PMLR.

\bibitem[{Kingma and Ba(2015)}]{kingma2015adam}
Diederik~P. Kingma and Jimmy Ba. 2015.
\newblock Adam: A method for stochastic optimization.
\newblock In \emph{Proceedings of the 3rd International Conference on Learning Representations (ICLR)}, San Diego, CA, USA.

\bibitem[{Li et~al.(2024)Li, Li, Song, Yang, Ma, and Yu}]{liPMETPreciseModel2024}
Xiaopeng Li, Shasha Li, Shezheng Song, Jing Yang, Jun Ma, and Jie Yu. 2024.
\newblock \href {https://doi.org/10.1609/aaai.v38i17.29818} {{PMET}: {Precise} {Model} {Editing} in a {Transformer}}.
\newblock \emph{Proceedings of the AAAI Conference on Artificial Intelligence}, 38(17):18564--18572.

\bibitem[{Liu et~al.(2025)Liu, Chen, Hirlimann, Hakimi, Wang, Kargaran, Rothe, Yvon, and Schuetze}]{liu-etal-2025-relation-specific}
Yihong Liu, Runsheng Chen, Lea Hirlimann, Ahmad~Dawar Hakimi, Mingyang Wang, Amir~Hossein Kargaran, Sascha Rothe, Fran{\c{c}}ois Yvon, and Hinrich Schuetze. 2025.
\newblock \href {https://doi.org/10.18653/v1/2025.emnlp-main.52} {On relation-specific neurons in large language models}.
\newblock In \emph{Proceedings of the 2025 Conference on Empirical Methods in Natural Language Processing}, pages 992--1022, Suzhou, China. Association for Computational Linguistics.

\bibitem[{Lv et~al.(2024)Lv, Chen, Zhang, Wang, Liu, Wen, Xie, and Yan}]{lvInterpretingKeyMechanisms2024}
Ang Lv, Yuhan Chen, Kaiyi Zhang, Yulong Wang, Lifeng Liu, Ji-Rong Wen, Jian Xie, and Rui Yan. 2024.
\newblock \href {https://doi.org/10.48550/arXiv.2403.19521} {Interpreting {Key} {Mechanisms} of {Factual} {Recall} in {Transformer}-{Based} {Language} {Models}}.
\newblock \emph{arXiv preprint}.
\newblock ArXiv:2403.19521 [cs].

\bibitem[{Meng et~al.(2022)Meng, Bau, Andonian, and Belinkov}]{meng2022locating}
Kevin Meng, David Bau, Alex Andonian, and Yonatan Belinkov. 2022.
\newblock \href {https://proceedings.neurips.cc/paper_files/paper/2022/file/6f1d43d5a82a37e89b0665b33bf3a182-Paper-Conference.pdf} {Locating and editing factual associations in gpt}.
\newblock In \emph{Advances in Neural Information Processing Systems}, volume~35, pages 17359--17372. Curran Associates, Inc.

\bibitem[{Morand et~al.(2025)Morand, Tomeh, Mothe, and Piwowarski}]{morandToMMeREfficientEntity2025}
Victor Morand, Nadi Tomeh, Josiane Mothe, and Benjamin Piwowarski. 2025.
\newblock \href {https://doi.org/10.48550/arXiv.2510.19410} {{ToMMeR} -- {Efficient} {Entity} {Mention} {Detection} from {Large} {Language} {Models}}.
\newblock \emph{arXiv preprint}.
\newblock ArXiv:2510.19410 [cs].

\bibitem[{Murdoch et~al.(2018)Murdoch, Liu, and Yu}]{james2018beyond}
W.~James Murdoch, Peter~J. Liu, and Bin Yu. 2018.
\newblock \href {https://openreview.net/forum?id=rkRwGg-0Z} {Beyond word importance: Contextual decomposition to extract interactions from {LSTM}s}.
\newblock In \emph{Proceedings of the 6th International Conference on Learning Representations (ICLR 2018)}.

\bibitem[{Popovic and F{\"a}rber(2024)}]{popovic-farber-2024-embedded}
Nicholas Popovic and Michael F{\"a}rber. 2024.
\newblock \href {https://doi.org/10.18653/v1/2024.emnlp-main.988} {Embedded named entity recognition using probing classifiers}.
\newblock In \emph{Proceedings of the 2024 Conference on Empirical Methods in Natural Language Processing}, pages 17830--17850, Miami, Florida, USA. Association for Computational Linguistics.

\bibitem[{Sakata et~al.(2025)Sakata, Heinzerling, Yokoi, Ito, and Inui}]{sakata-etal-2025-entity}
Masaki Sakata, Benjamin Heinzerling, Sho Yokoi, Takumi Ito, and Kentaro Inui. 2025.
\newblock \href {https://doi.org/10.18653/v1/2025.findings-acl.858} {On entity identification in language models}.
\newblock In \emph{Findings of the Association for Computational Linguistics: ACL 2025}, pages 16717--16741, Vienna, Austria. Association for Computational Linguistics.

\bibitem[{Sundararajan et~al.(2017)Sundararajan, Taly, and Yan}]{pmlr-v70-sundararajan17a}
Mukund Sundararajan, Ankur Taly, and Qiqi Yan. 2017.
\newblock \href {https://proceedings.mlr.press/v70/sundararajan17a.html} {Axiomatic attribution for deep networks}.
\newblock In \emph{Proceedings of the 34th International Conference on Machine Learning}, volume~70 of \emph{Proceedings of Machine Learning Research}, pages 3319--3328. PMLR.

\bibitem[{Todd et~al.(2024)Todd, Li, Sharma, Mueller, Wallace, and Bau}]{todd2024function}
Eric Todd, Millicent Li, Arnab~Sen Sharma, Aaron Mueller, Byron~C Wallace, and David Bau. 2024.
\newblock \href {https://openreview.net/forum?id=AwyxtyMwaG} {Function vectors in large language models}.
\newblock In \emph{The Twelfth International Conference on Learning Representations}.

\bibitem[{Wang et~al.(2025)Wang, Wan, Hu, Zhang, Tian, Chen, Shen, and Ye}]{wang-etal-2025-tracing}
Yiqun Wang, Chaoqun Wan, Sile Hu, Yonggang Zhang, Xiang Tian, Yaowu Chen, Xu~Shen, and Jieping Ye. 2025.
\newblock \href {https://doi.org/10.18653/v1/2025.acl-long.1133} {Tracing and dissecting how {LLM}s recall factual knowledge for real world questions}.
\newblock In \emph{Proceedings of the 63rd Annual Meeting of the Association for Computational Linguistics (Volume 1: Long Papers)}, pages 23246--23271, Vienna, Austria. Association for Computational Linguistics.

\bibitem[{Wang et~al.(2024)Wang, Whyte, and Xu}]{wang-etal-2024-locating}
Zijian Wang, Britney Whyte, and Chang Xu. 2024.
\newblock \href {https://doi.org/10.18653/v1/2024.findings-acl.287} {Locating and extracting relational concepts in large language models}.
\newblock In \emph{Findings of the Association for Computational Linguistics: ACL 2024}, pages 4818--4832, Bangkok, Thailand. Association for Computational Linguistics.

\bibitem[{Yang et~al.(2025)Yang, Li, Yang, Zhang, Hui, Zheng, Yu, Gao, Huang, Lv, Zheng, Liu, Zhou, Huang, Hu, Ge, Wei, Lin, Tang, Yang, Tu, Zhang, Yang, Yang, Zhou, Zhou, Lin, Dang, Bao, Yang, Yu, Deng, Li, Xue, Li, Zhang, Wang, Zhu, Men, Gao, Liu, Luo, Li, Tang, Yin, Ren, Wang, Zhang, Ren, Fan, Su, Zhang, Zhang, Wan, Liu, Wang, Cui, Zhang, Zhou, and Qiu}]{qwen3}
An~Yang, Anfeng Li, Baosong Yang, Beichen Zhang, Binyuan Hui, Bo~Zheng, Bowen Yu, Chang Gao, Chengen Huang, Chenxu Lv, Chujie Zheng, Dayiheng Liu, Fan Zhou, Fei Huang, Feng Hu, Hao Ge, Haoran Wei, Huan Lin, Jialong Tang, and 41 others. 2025.
\newblock Qwen3 technical report.
\newblock \emph{arXiv preprint arXiv:2505.09388}.

\bibitem[{Yu and Ananiadou(2024)}]{yu-ananiadou-2024-neuron}
Zeping Yu and Sophia Ananiadou. 2024.
\newblock \href {https://doi.org/10.18653/v1/2024.emnlp-main.191} {Neuron-level knowledge attribution in large language models}.
\newblock In \emph{Proceedings of the 2024 Conference on Empirical Methods in Natural Language Processing}, pages 3267--3280, Miami, Florida, USA. Association for Computational Linguistics.

\bibitem[{Yu et~al.(2025)Yu, Belinkov, and Ananiadou}]{yu-etal-2025-back}
Zeping Yu, Yonatan Belinkov, and Sophia Ananiadou. 2025.
\newblock \href {https://doi.org/10.18653/v1/2025.emnlp-main.567} {Back attention: Understanding and enhancing multi-hop reasoning in large language models}.
\newblock In \emph{Proceedings of the 2025 Conference on Empirical Methods in Natural Language Processing}, pages 11268--11283, Suzhou, China. Association for Computational Linguistics.

\bibitem[{Zheng et~al.(2025)Zheng, Wang, Huang, Song, Yang, Tang, Xiong, and Li}]{attentionheadsurvey2024}
Zifan Zheng, Yezhaohui Wang, Yuxin Huang, Shichao Song, Mingchuan Yang, Bo~Tang, Feiyu Xiong, and Zhiyu Li. 2025.
\newblock \href {https://doi.org/10.1016/j.patter.2025.101176} {Attention heads of large language models}.
\newblock \emph{Patterns}, 6(2):101176.

\end{thebibliography}

\appendix

\section{Few-shot RelSpec Feature Selection}
\label{sec:appendix-RelSpec-fewshot}

In the few-shot experiments, we select relation-specific features per episode
using the same average-precision (AP) ranking idea as
\citet{liu-etal-2025-relation-specific}, computed only on the episode
support set.

\paragraph{Episode setup.}
For each episode, we sample $n$ relation classes. For every class we sample
$k$ support examples and $q$ query examples.

\paragraph{Expertise scores (AP).}
Given a fixed episode and a fixed class $c$, we treat support examples of
$c$ as positives and the other support examples as negatives. For every
feature in the chosen relational knowledge recall position representation, we rank support examples by that
feature's activation value and compute average precision (AP) for this
binary classification. This gives an AP score per feature for class $c$.

\paragraph{Selecting RelSpec Features.}
For each class $c$ in the episode, we take the top $m$ features by AP
(default $m=3000$). If the same feature appears in the top-$m$ lists of
multiple classes, we optionally deduplicate it within the episode.

\paragraph{Probe training.}
We build feature vectors by reading out the selected features for every
support/query example, train a linear classifier on the support set (cross
entropy, Adam optimizer \cite{kingma2015adam}, 200 epochs), and evaluate on the query set. Results are averaged over multiple episodes and random seeds.
\onecolumn

\section{Illustration of Probing Locations}
\label{sec:appendix-probing-illustrations}
\begin{figure}[!ht]
    \centering
    \includegraphics[width=0.4\linewidth]{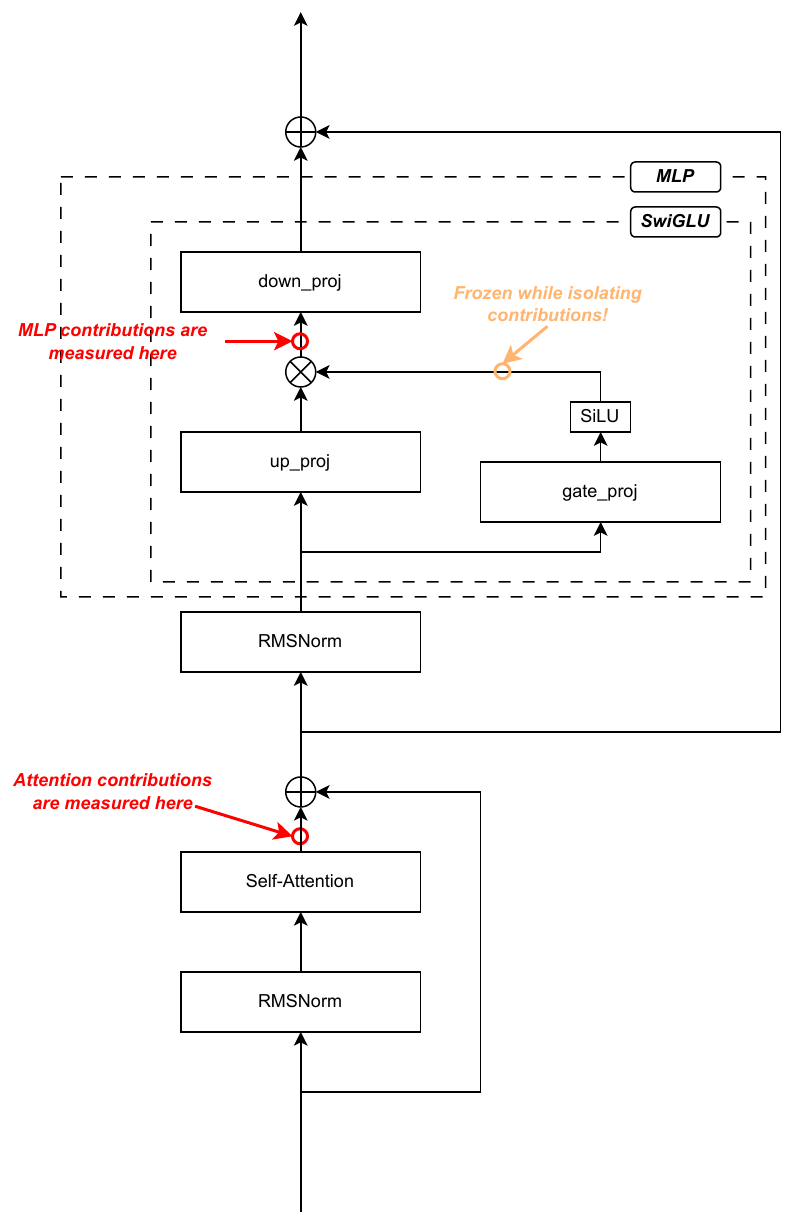}
    \caption{Overview of where attention and MLP contributions are measured in the LLaMA architecture. The main information gain by measuring MLP contributions over attention contributions is that they incorporate the non-linearity introduced by SiGLU, which is why we measure right after that. Measuring before down\_proj results in higher sparsity. In order for the individual contributions to be isolated correctly, we freeze the gate post linearity.}
    \label{fig:llamablock}
\end{figure}

Propagating an isolated per-head attention contribution through the MLP requires handling RMSNorm and the gated non-linearity of SiGLU. We adopt a first-order linearization around the full-pass sublayer input: the RMSNorm scale factors, the gate activations $\sigma(W_\text{gate}\,x_t)$, and the up-projection pre-activations $W_\text{up}\,x_t$ computed during the regular forward pass at the target token $t$ are cached and reused when propagating each per-head contribution $\Delta_{\text{att},h}(t)$. The resulting $\Delta_{\text{MLP},h}$ is therefore exact in the linear components (RMSNorm scaling and $W_\text{up}$) and first-order in the gate, so summing $\Delta_{\text{MLP},h}$ over $h$ approximates but does not exactly reproduce the full MLP sublayer output.

\newpage
\FloatBarrier
\section{$16$-way-$5$-shot Classification Reports}
\label{sec:appendix-16way}

\begin{table}[h]
\centering
\scriptsize
\setlength{\tabcolsep}{4pt}
\resizebox{\linewidth}{!}{%
\begin{tabular}{lrrrccc}
\hline
\textbf{Property} & \textbf{Output range} & \textbf{Avg. conn.} & \textbf{TF-IDF sim. ($\times 10^{-3}$)} & \textbf{Precision (\%)} & \textbf{Recall (\%)} & \textbf{F1 (\%)} \\
\hline
P59\phantom{00} - constellation & 88 & 1.000 & 31.8 & 99.66 & 98.20 & 98.93 \\
P177\phantom{0} - crosses & 15,572 & 1.085 & 48.4 & 88.19 & 91.13 & 89.64 \\
P2094 - competition class & 1,298 & 1.005 & 24.6 & 89.58 & 87.67 & 88.61 \\
P364\phantom{0} - original language of film or TV show & 783 & 1.001 & 49.8 & 84.52 & 89.20 & 86.80 \\
P412\phantom{0} - voice type & 96 & 1.007 & 28.3 & 83.57 & 82.40 & 82.98 \\
P413\phantom{0} - position played on team / speciality & 408 & 1.005 & 30.0 & 82.88 & 80.07 & 81.45 \\
P206\phantom{0} - located in or next to body of water & 23,979 & 1.096 & 0.0 & 74.88 & 81.87 & 78.22 \\
P410\phantom{0} - military, police or special rank & 1,908 & 1.001 & 28.7 & 68.22 & 76.27 & 72.02 \\
P641\phantom{0} - sport & 2,170 & 1.001 & 6.9 & 71.23 & 69.00 & 70.10 \\
P155\phantom{0} - follows & 1,706,612 & 1.935 & 0.0 & 63.37 & 65.40 & 64.37 \\
P921\phantom{0} - main subject & 524,961 & 1.991 & 0.0 & 63.79 & 56.60 & 59.98 \\
P463\phantom{0} - member of & 79,764 & 1.097 & 0.0 & 60.79 & 56.53 & 58.58 \\
P25\phantom{00} - mother & 371,609 & 2.000 & 0.0 & 39.99 & 43.13 & 41.50 \\
P40\phantom{00} - child & 1,216,423 & 1.998 & 0.0 & 41.14 & 39.80 & 40.46 \\
P26\phantom{00} - spouse & 826,642 & 2.030 & 0.0 & 38.01 & 39.93 & 38.95 \\
P361\phantom{0} - part of & 576,931 & 1.660 & 0.0 & 38.57 & 33.53 & 35.88 \\
\hline
\textbf{Macro avg} &  &  &  & 68.02 & 68.17 & 68.03 \\
\textbf{Accuracy} &   &  &  &  &  & 68.17 \\
\hline
\end{tabular}
}
\caption{Per-property precision, recall, F1-score, macro averages, and accuracy (percentages) for LLaMA-3.2$_\textnormal{\small{1B-Instruct}}$ in the $16$-way-$5$-shot order-fix setting, sorted by per-property F1. Output range is the number of distinct Wikidata objects for a property; avg.\ conn.\ is the mean number of direct truthy properties connecting $1000$ sampled distinct subject--object pairs for which that property holds; TF-IDF sim.\ is the median pairwise cosine similarity between TF-IDF representations of raw FewRel examples for that relation, reported in units of $10^{-3}$.}
\label{tab:appendix-16way-llama32-1b}
\end{table}

\begin{table}[h]
\centering
\scriptsize
\setlength{\tabcolsep}{4pt}
\resizebox{\linewidth}{!}{%
\begin{tabular}{lrrrccc}
\hline
\textbf{Property} & \textbf{Output range} & \textbf{Avg. conn.} & \textbf{TF-IDF sim. ($\times 10^{-3}$)} & \textbf{Precision (\%)} & \textbf{Recall (\%)} & \textbf{F1 (\%)} \\
\hline
P59\phantom{00} - constellation & 88 & 1.000 & 31.8 & 98.35 & 99.07 & 98.70 \\
P2094 - competition class & 1,298 & 1.005 & 24.6 & 95.57 & 89.20 & 92.28 \\
P177\phantom{0} - crosses & 15,572 & 1.085 & 48.4 & 89.09 & 89.80 & 89.44 \\
P364\phantom{0} - original language of film or TV show & 783 & 1.001 & 49.8 & 90.58 & 87.13 & 88.82 \\
P412\phantom{0} - voice type & 96 & 1.007 & 28.3 & 90.71 & 85.93 & 88.26 \\
P413\phantom{0} - position played on team / speciality & 408 & 1.005 & 30.0 & 86.52 & 79.60 & 82.92 \\
P206\phantom{0} - located in or next to body of water & 23,979 & 1.096 & 0.0 & 74.56 & 79.33 & 76.87 \\
P410\phantom{0} - military, police or special rank & 1,908 & 1.001 & 28.7 & 72.97 & 80.27 & 76.44 \\
P641\phantom{0} - sport & 2,170 & 1.001 & 6.9 & 73.81 & 72.13 & 72.96 \\
P155\phantom{0} - follows & 1,706,612 & 1.935 & 0.0 & 60.29 & 68.33 & 64.06 \\
P463\phantom{0} - member of & 79,764 & 1.097 & 0.0 & 66.16 & 55.13 & 60.15 \\
P921\phantom{0} - main subject & 524,961 & 1.991 & 0.0 & 65.57 & 54.33 & 59.42 \\
P40\phantom{00} - child & 1,216,423 & 1.998 & 0.0 & 47.07 & 46.13 & 46.60 \\
P25\phantom{00} - mother & 371,609 & 2.000 & 0.0 & 42.70 & 47.80 & 45.11 \\
P26\phantom{00} - spouse & 826,642 & 2.030 & 0.0 & 39.15 & 39.47 & 39.31 \\
P361\phantom{0} - part of & 576,931 & 1.660 & 0.0 & 35.94 & 42.33 & 38.87 \\
\hline
\textbf{Macro avg} &  &  &  & 70.56 & 69.75 & 70.01 \\
\textbf{Accuracy} &   &  &  &  &  & 69.75 \\
\hline
\end{tabular}
}
\caption{Per-property precision, recall, F1-score, macro averages, and accuracy (percentages) for LLaMA-3.2$_\textnormal{\small{3B-Instruct}}$ in the $16$-way-$5$-shot order-fix setting, sorted by per-property F1. Output range is the number of distinct Wikidata objects for a property; avg.\ conn.\ is the mean number of direct truthy properties connecting $1000$ sampled distinct subject--object pairs for which that property holds; TF-IDF sim.\ is the median pairwise cosine similarity between TF-IDF representations of raw FewRel examples for that relation, reported in units of $10^{-3}$.}
\label{tab:appendix-16way-llama32-3b}
\end{table}

\begin{table}[h]
\centering
\scriptsize
\setlength{\tabcolsep}{4pt}
\resizebox{\linewidth}{!}{%
\begin{tabular}{lrrrccc}
\hline
\textbf{Property} & \textbf{Output range} & \textbf{Avg. conn.} & \textbf{TF-IDF sim. ($\times 10^{-3}$)} & \textbf{Precision (\%)} & \textbf{Recall (\%)} & \textbf{F1 (\%)} \\
\hline
P59\phantom{00} - constellation & 88 & 1.000 & 31.8 & 98.81 & 99.67 & 99.24 \\
P2094 - competition class & 1,298 & 1.005 & 24.6 & 97.82 & 89.60 & 93.53 \\
P412\phantom{0} - voice type & 96 & 1.007 & 28.3 & 94.19 & 87.53 & 90.74 \\
P364\phantom{0} - original language of film or TV show & 783 & 1.001 & 49.8 & 90.01 & 90.67 & 90.34 \\
P177\phantom{0} - crosses & 15,572 & 1.085 & 48.4 & 87.93 & 91.33 & 89.60 \\
P413\phantom{0} - position played on team / speciality & 408 & 1.005 & 30.0 & 85.59 & 83.53 & 84.55 \\
P410\phantom{0} - military, police or special rank & 1,908 & 1.001 & 28.7 & 77.15 & 85.07 & 80.91 \\
P206\phantom{0} - located in or next to body of water & 23,979 & 1.096 & 0.0 & 75.54 & 82.13 & 78.70 \\
P641\phantom{0} - sport & 2,170 & 1.001 & 6.9 & 76.22 & 76.07 & 76.14 \\
P155\phantom{0} - follows & 1,706,612 & 1.935 & 0.0 & 65.88 & 74.27 & 69.82 \\
P463\phantom{0} - member of & 79,764 & 1.097 & 0.0 & 68.75 & 60.00 & 64.08 \\
P921\phantom{0} - main subject & 524,961 & 1.991 & 0.0 & 69.62 & 55.60 & 61.82 \\
P40\phantom{00} - child & 1,216,423 & 1.998 & 0.0 & 47.27 & 47.93 & 47.60 \\
P25\phantom{00} - mother & 371,609 & 2.000 & 0.0 & 43.89 & 49.53 & 46.54 \\
P26\phantom{00} - spouse & 826,642 & 2.030 & 0.0 & 40.09 & 41.27 & 40.67 \\
P361\phantom{0} - part of & 576,931 & 1.660 & 0.0 & 40.47 & 39.07 & 39.76 \\
\hline
\textbf{Macro avg} &  &  &  & 72.45 & 72.08 & 72.13 \\
\textbf{Accuracy} &   &  &  &  &  & 72.08 \\
\hline
\end{tabular}
}
\caption{Per-property precision, recall, F1-score, macro averages, and accuracy (percentages) for LLaMA-3.1$_\textnormal{\small{8B-Instruct}}$ in the $16$-way-$5$-shot order-fix setting, sorted by per-property F1. Output range is the number of distinct Wikidata objects for a property; avg.\ conn.\ is the mean number of direct truthy properties connecting $1000$ sampled distinct subject--object pairs for which that property holds; TF-IDF sim.\ is the median pairwise cosine similarity between TF-IDF representations of raw FewRel examples for that relation, reported in units of $10^{-3}$.}
\label{tab:appendix-16way-llama31-8b}
\end{table}

\begin{table}[h]
\centering
\scriptsize
\setlength{\tabcolsep}{4pt}
\resizebox{\linewidth}{!}{%
\begin{tabular}{lrrrccc}
\hline
\textbf{Property} & \textbf{Output range} & \textbf{Avg. conn.} & \textbf{TF-IDF sim. ($\times 10^{-3}$)} & \textbf{Precision (\%)} & \textbf{Recall (\%)} & \textbf{F1 (\%)} \\
\hline
P59\phantom{00} - constellation & 88 & 1.000 & 31.8 & 98.87 & 98.73 & 98.80 \\
P2094 - competition class & 1,298 & 1.005 & 24.6 & 82.97 & 95.13 & 88.63 \\
P413\phantom{0} - position played on team / speciality & 408 & 1.005 & 30.0 & 79.96 & 87.80 & 83.70 \\
P412\phantom{0} - voice type & 96 & 1.007 & 28.3 & 78.08 & 88.60 & 83.01 \\
P364\phantom{0} - original language of film or TV show & 783 & 1.001 & 49.8 & 73.98 & 94.20 & 82.87 \\
P177\phantom{0} - crosses & 15,572 & 1.085 & 48.4 & 76.77 & 89.00 & 82.43 \\
P641\phantom{0} - sport & 2,170 & 1.001 & 6.9 & 73.21 & 76.53 & 74.84 \\
P410\phantom{0} - military, police or special rank & 1,908 & 1.001 & 28.7 & 69.68 & 79.67 & 74.34 \\
P206\phantom{0} - located in or next to body of water & 23,979 & 1.096 & 0.0 & 62.08 & 73.47 & 67.30 \\
P155\phantom{0} - follows & 1,706,612 & 1.935 & 0.0 & 76.52 & 58.87 & 66.54 \\
P463\phantom{0} - member of & 79,764 & 1.097 & 0.0 & 68.59 & 49.07 & 57.21 \\
P921\phantom{0} - main subject & 524,961 & 1.991 & 0.0 & 57.36 & 52.20 & 54.66 \\
P40\phantom{00} - child & 1,216,423 & 1.998 & 0.0 & 47.56 & 44.27 & 45.86 \\
P25\phantom{00} - mother & 371,609 & 2.000 & 0.0 & 43.94 & 44.20 & 44.07 \\
P26\phantom{00} - spouse & 826,642 & 2.030 & 0.0 & 37.43 & 38.20 & 37.81 \\
P361\phantom{0} - part of & 576,931 & 1.660 & 0.0 & 42.41 & 19.73 & 26.93 \\
\hline
\textbf{Macro avg} &  &  &  & 66.84 & 68.10 & 66.81 \\
\textbf{Accuracy} &   &  &  &  &  & 68.10 \\
\hline
\end{tabular}
}
\caption{Per-property precision, recall, F1-score, macro averages, and accuracy (percentages) for Qwen3$_\textnormal{\small{4B-Instruct}}$ in the $16$-way-$5$-shot order-fix setting, sorted by per-property F1. Output range is the number of distinct Wikidata objects for a property; avg.\ conn.\ is the mean number of direct truthy properties connecting $1000$ sampled distinct subject--object pairs for which that property holds; TF-IDF sim.\ is the median pairwise cosine similarity between TF-IDF representations of raw FewRel examples for that relation, reported in units of $10^{-3}$.}
\label{tab:appendix-16way-qwen3-4b}
\end{table}

\newpage
\FloatBarrier
\section{Few-Shot Performance for Varying Amounts of RelSpec Features}
\label{sec:appendix-num-RelSpec}

\begin{table}[H]
    \centering
    \begin{tabular}{l|cccc}
        Model & 5 way 1 shot & 5 way 5 shot & 10 way 1 shot & 10 way 5 shot \\
        \hline
        \multicolumn{5}{c}{feature size $200$} \\
        \hline
        LLaMA-3.2$_\textnormal{\small{1B-Instruct}}$ & $36.15 \pm 0.39\%$ & $86.63 \pm 0.27\%$ & $25.99 \pm 0.36\%$ & $82.43 \pm 0.23\%$ \\
        LLaMA-3.2$_\textnormal{\small{3B-Instruct}}$ & $34.91 \pm 0.25\%$ & $86.13 \pm 0.16\%$ & $24.86 \pm 0.38\%$ & $83.31 \pm 0.21\%$ \\
        LLaMA-3.1$_\textnormal{\small{8B-Instruct}}$ & $43.43 \pm 0.20\%$ & $84.03 \pm 0.33\%$ & $30.83 \pm 0.30\%$ & $83.37 \pm 0.11\%$ \\
        \hline
        \multicolumn{5}{c}{feature size $1000$} \\
        \hline
        LLaMA-3.2$_\textnormal{\small{1B-Instruct}}$ & $39.51 \pm 0.25\%$ & $89.69 \pm 0.20\%$ & $28.44 \pm 0.49\%$ & $83.07 \pm 0.21\%$ \\
        LLaMA-3.2$_\textnormal{\small{3B-Instruct}}$ & $37.17 \pm 0.22\%$ & $90.30 \pm 0.15\%$ & $26.74 \pm 0.38\%$ & $84.34 \pm 0.17\%$ \\
        LLaMA-3.1$_\textnormal{\small{8B-Instruct}}$ & $43.61 \pm 0.23\%$ & $89.72 \pm 0.17\%$ & $31.17 \pm 0.47\%$ & $84.87 \pm 0.17\%$ \\
        \hline
        \multicolumn{5}{c}{feature size $3000$} \\
        \hline
        LLaMA-3.2$_\textnormal{\small{1B-Instruct}}$ & $38.94 \pm 0.29\%$ & $90.26 \pm 0.19\%$ & $31.93 \pm 0.41\%$ & $82.82 \pm 0.23\%$ \\
        LLaMA-3.2$_\textnormal{\small{3B-Instruct}}$ & $37.64 \pm 0.20 \%$ & $91.06 \pm 0.16 \%$ & $30.89 \pm 0.40 \%$ & $83.91 \pm 0.15 \%$ \\
        LLaMA-3.1$_\textnormal{\small{8B-Instruct}}$ & $42.75 \pm 0.22\%$ & $91.09 \pm 0.17\%$ & $30.78 \pm 0.45\%$ & $84.71 \pm 0.13\%$ \\
        \hline
        \multicolumn{5}{c}{feature size $5000$} \\
        \hline
        LLaMA-3.2$_\textnormal{\small{1B-Instruct}}$ & $41.89 \pm 0.21\%$ & $90.12 \pm 0.19\%$ & $32.57 \pm 0.45\%$ & $82.12 \pm 0.21\%$ \\
        LLaMA-3.2$_\textnormal{\small{3B-Instruct}}$ & $40.98 \pm 0.22\%$ & $90.92 \pm 0.16\%$ & $31.98 \pm 0.48\%$ & $83.04 \pm 0.17\%$ \\
        LLaMA-3.1$_\textnormal{\small{8B-Instruct}}$ & $42.57 \pm 0.13\%$ & $91.25 \pm 0.17\%$ & $34.36 \pm 0.45\%$ & $84.01 \pm 0.12\%$ \\
        \hline

    \end{tabular}
    \caption{Accuracy of probes for $\Delta_{\text{att},h}$ on FewRel for 500 episodes, using different amounts of RelSpec features and 15 queries per class, per episode, averaged across 5 random seeds.}
    \label{tab:RelSpecvariation}
\end{table}

\newpage
\FloatBarrier
\section{Detailed Results for Relation Classification}
\label{sec:appendix-fewshot}
See Tables \ref{tab:5-way-1-shot-m-3000},\ref{tab:5-way-5-shot-m-3000},\ref{tab:10-way-1-shot-m-3000}, and \ref{tab:10-way-5-shot-m-3000}.
\nopagebreak[4]
\FloatBarrier

\begin{table}[H]
    \centering
    \resizebox{0.75\linewidth}{!}{
    \begin{tabular}{l|cccc}
         & \multicolumn{3}{c|}{LLaMA-3.2} & Qwen-3 \\
        Feature & 1B & 3B & 8B & 4B \\
        \hline
        MLP & $70.85 \pm 0.32\%$ & $41.71 \pm 0.20\%$ & $38.35 \pm 0.23\%$ & $40.15 \pm 0.20\%$ \\
        $\Delta_{\text{MLP},h}$  & $40.01 \pm 0.20\%$ & $37.92 \pm 0.22\%$ & $40.14 \pm 0.18\%$ & $39.74 \pm 0.13\%$ \\
        $\Delta_{\text{MLP},e_1}$ & $38.93 \pm 0.20\%$ & $28.88 \pm 0.10\%$ & $27.43 \pm 0.13\%$ & $33.40 \pm 0.17\%$ \\
        $\Delta_{\text{MLP},e_1,h}$ & $26.54 \pm 0.33\%$ & $27.24 \pm 0.16\%$ & $26.51 \pm 0.13\%$ & $28.28 \pm 0.26\%$ \\
        \hline
        Attention & $59.94 \pm 0.57\%$ & $65.94 \pm 0.49\%$ & $58.93 \pm 0.68\%$ & $58.62 \pm 0.27\%$ \\
        $\Delta_{\text{att},h}$ & $38.94 \pm 0.29\%$ & $37.64 \pm 0.20\%$ & $42.75 \pm 0.22\%$ & $40.66 \pm 0.19\%$ \\
        $\Delta_{\text{att},e_1}$ & $37.11 \pm 0.36\%$ & $35.76 \pm 0.14\%$ & $35.31 \pm 0.25\%$ & $36.95 \pm 0.30\%$ \\
        $\Delta_{\text{att},e_1,h}$ & $28.44 \pm 0.38\%$ & $28.56 \pm 0.12\%$ & $28.82 \pm 0.26\%$ & $29.47 \pm 0.19\%$ \\
        \hline
    \end{tabular}
    }
    \caption{Few-shot relation classification accuracy on FewRel validation split for $n=5,k=1$ with constraint $e_1=s$ and $e_2=o$.}
    \label{tab:5-way-1-shot-m-3000}
    \centering
    \resizebox{0.75\linewidth}{!}{
    \begin{tabular}{l|cccc}
         & \multicolumn{3}{c|}{LLaMA-3.2} & Qwen-3  \\
        Feature & 1B & 3B & 8B & 4B  \\
        \hline
        MLP & $85.79 \pm 0.15\%$ & $86.40 \pm 0.17\%$ & $85.90 \pm 0.22\%$ & $80.37 \pm 0.23\%$ \\
        $\Delta_{\text{MLP},h}$  & $89.96 \pm 0.20\%$ & $90.90 \pm 0.12\%$ & $89.99 \pm 0.15 \%$ & $88.43 \pm 0.25\%$ \\
        $\Delta_{\text{MLP},e_1}$ & $66.66 \pm 0.26\%$ & $66.02 \pm 0.41\%$ & $65.85 \pm 0.16\%$ & $68.49 \pm 0.30\%$ \\
        $\Delta_{\text{MLP},e_1,h}$ & $72.56 \pm 0.29\%$ & $72.33 \pm 0.30\%$ & $71.59 \pm 0.13\%$ & $73.98 \pm 0.23\%$ \\        
        \hline
        Attention & $83.65 \pm 0.16\%$ & $86.61 \pm 0.16\%$ & $86.79 \pm 0.08\%$ & $75.06 \pm 0.19\%$ \\
        $\Delta_{\text{att},h}$ & $90.26 \pm 0.19\%$ & $91.06 \pm 0.16 \%$ & $91.09 \pm 0.17\%$ & $89.66 \pm 0.19\%$ \\
        $\Delta_{\text{att},e_1}$ & $59.64 \pm 0.36\%$ & $60.16 \pm 0.42\%$ & $59.85 \pm 0.33\%$ & $59.58 \pm 0.23\%$ \\
        $\Delta_{\text{att},e_1,h}$ & $64.96 \pm 0.39\%$ & $65.21 \pm 0.45\%$ & $64.59 \pm 0.43\%$ & $65.29 \pm 0.39\%$ \\   
        \hline
    \end{tabular}
    }
    \caption{Few-shot relation classification accuracy on FewRel validation split for $n=5,k=5$ with constraint $e_1=s$ and $e_2=o$.}
    \label{tab:5-way-5-shot-m-3000}
    \centering
    \resizebox{0.75\linewidth}{!}{
    \begin{tabular}{l|cccc}
         & \multicolumn{3}{c|}{LLaMA-3.2} & Qwen-3 \\
        Feature & 1B & 3B & 8B & 4B \\
        \hline
        MLP & $60.93 \pm 0.25\%$ & $38.09 \pm 0.35\%$ & $32.90 \pm 0.30\%$ & $31.04 \pm 0.43\%$ \\
        $\Delta_{\text{MLP},h}$  & $29.78 \pm 0.38\%$ & $28.52 \pm 0.32\%$ & $30.66 \pm 0.41\%$ & $30.82 \pm 0.38\%$ \\
        $\Delta_{\text{MLP},e_1}$ & $31.29 \pm 0.11\%$ & $25.58 \pm 0.27\%$ & $21.65 \pm 0.28\%$ & $24.66 \pm 0.33\%$ \\
        $\Delta_{\text{MLP},e_1,h}$ & $18.41 \pm 0.25\%$ & $18.87 \pm 0.21\%$ & $18.39 \pm 0.16\%$ & $20.93 \pm 0.22\%$ \\
        \hline
        Attention & $53.83 \pm 0.21\%$ & $58.39 \pm 0.20\%$ & $58.81 \pm 0.25\%$ & $45.94 \pm 0.42\%$ \\
        $\Delta_{\text{att},h}$ & $31.93 \pm 0.41\%$ & $30.89 \pm 0.40\%$ & $30.78 \pm 0.45\%$ & $31.48 \pm 0.33\%$ \\
        $\Delta_{\text{att},e_1}$ & $27.26 \pm 0.19\%$ & $26.51 \pm 0.07\%$ & $25.77 \pm 0.14\%$ & $27.26 \pm 0.16\%$ \\
        $\Delta_{\text{att},e_1,h}$ & $19.31 \pm 0.17\%$ & $19.54 \pm 0.18\%$ & $19.48 \pm 0.20\%$ & $19.38 \pm 0.18\%$ \\
        \hline
    \end{tabular}
    }
    \caption{Few-shot relation classification accuracy on FewRel validation split for $n=10,k=1$ with constraint $e_1=s$ and $e_2=o$.}
    \label{tab:10-way-1-shot-m-3000}
    \centering
    \resizebox{0.75\linewidth}{!}{
    \begin{tabular}{l|cccc}
         & \multicolumn{3}{c|}{LLaMA-3.2} & Qwen-3 \\
        Feature & 1B & 3B & 8B & 4B \\
        \hline
        MLP & $76.89 \pm 0.30\%$ & $78.35 \pm 0.19\%$ & $78.40 \pm 0.15\%$ & $70.68 \pm 0.13\%$ \\
        $\Delta_{\text{MLP},h}$  & $83.38 \pm 0.19\%$ & $85.13 \pm 0.15\%$ & $85.84 \pm 0.15\%$ & $84.09 \pm 0.18\%$ \\
        $\Delta_{\text{MLP},e_1}$ & $57.60 \pm 0.17\%$ & $56.70 \pm 0.13\%$ & $56.72 \pm 0.08\%$ & $58.24 \pm 0.18\%$ \\
        $\Delta_{\text{MLP},e_1,h}$ & $64.36 \pm 0.12\%$ & $64.38 \pm 0.12\%$ & $64.64 \pm 0.18\%$ & $66.65 \pm 0.15\%$ \\
        \hline
        Attention & $70.13 \pm 0.31\%$ & $74.25 \pm 0.26\%$ & $72.99 \pm 0.24\%$ & $63.95 \pm 0.30\%$ \\
        $\Delta_{\text{att},h}$ & $82.82 \pm 0.23\%$ & $83.91 \pm 0.15\%$ & $84.71 \pm 0.13\%$ & $80.88 \pm 0.12\%$ \\
        $\Delta_{\text{att},e_1}$ & $48.82 \pm 0.18\%$ & $49.08 \pm 0.23\%$ & $48.92 \pm 0.15\%$ & $48.05 \pm 0.18\%$ \\
        $\Delta_{\text{att},e_1,h}$ & $53.64 \pm 0.23\%$ & $54.19 \pm 0.24\%$ & $53.88 \pm 0.13\%$ & $53.25 \pm 0.16\%$ \\
        \hline
    \end{tabular}
    }
    \caption{Few-shot relation classification accuracy on FewRel validation split for $n=10,k=5$ with constraint $e_1=s$ and $e_2=o$.}
    \label{tab:10-way-5-shot-m-3000}
\end{table}

\FloatBarrier
\newpage

\section{Additional fill-in-the-blanks examples}
\label{section:appendix-fill-in-blanks-examples}
\begin{figure}[H]
  \centering

  \begin{subfigure}{\textwidth}
    \centering
    \includegraphics[width=\linewidth]{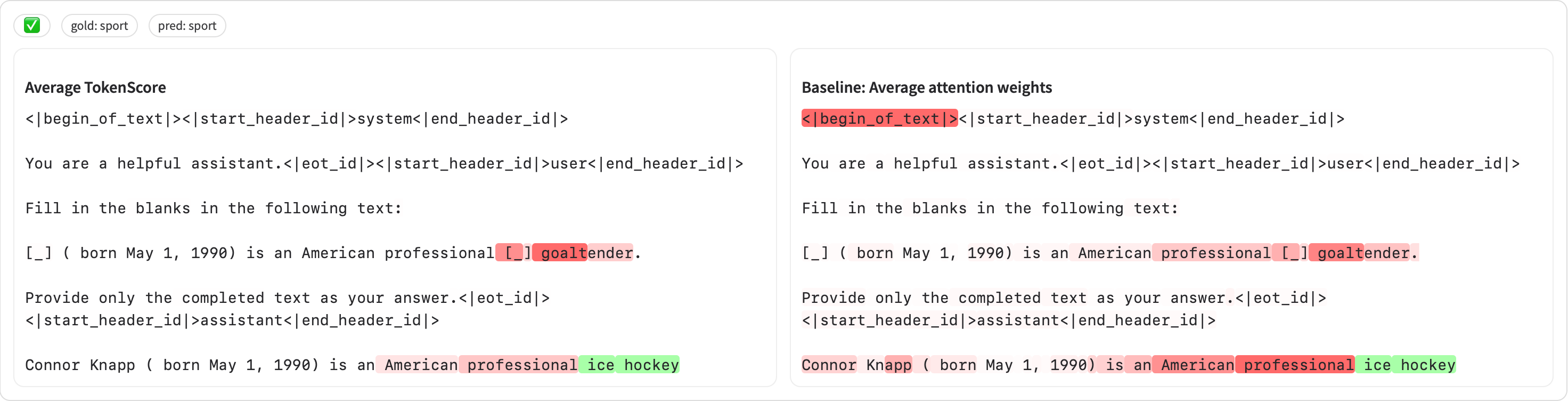}
  \end{subfigure}
  \begin{subfigure}{\textwidth}
    \centering
    \includegraphics[width=\linewidth]{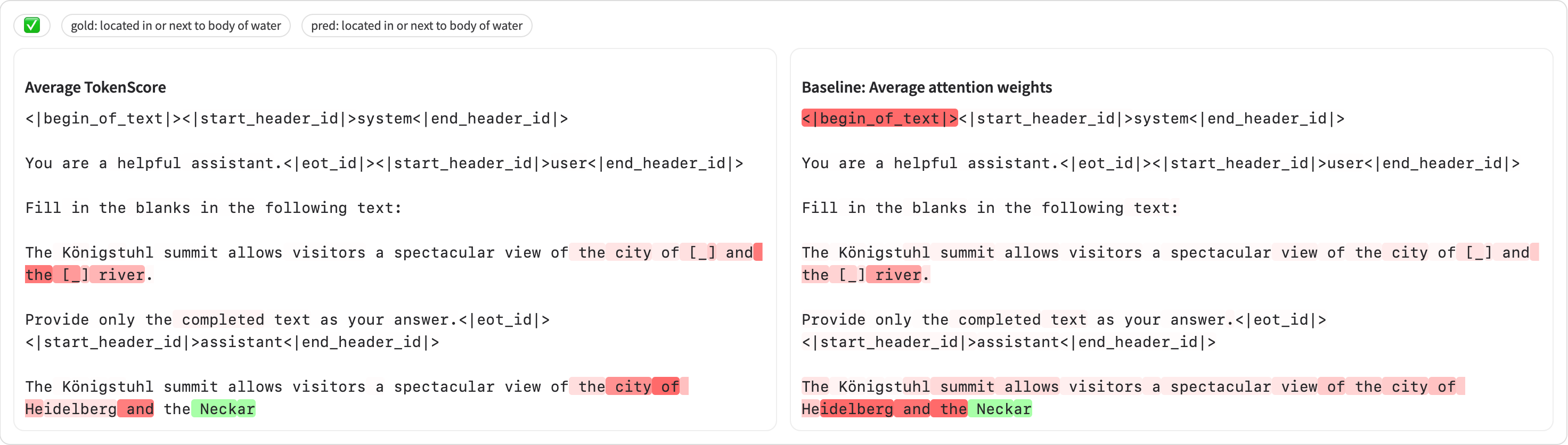}
  \end{subfigure}
  \begin{subfigure}{\textwidth}
    \centering
    \includegraphics[width=\linewidth]{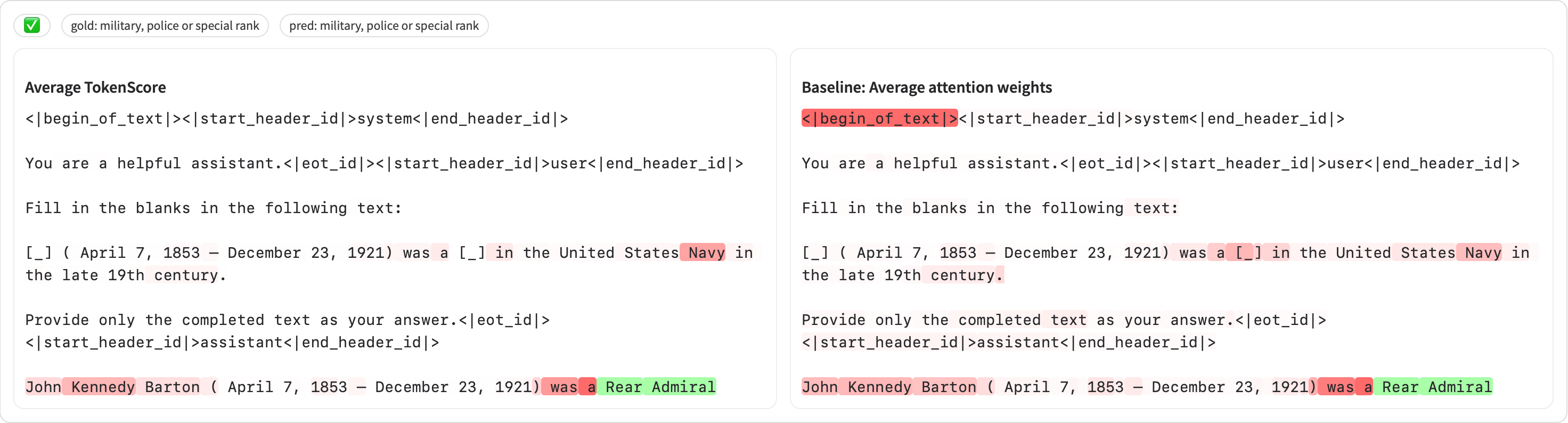}
  \end{subfigure}\

  \caption{Additional examples showing relevant information after tail entity mention, available due to fill-in-the-blanks format. Shown are both the aggregated TokenScore, as well as the average attention weights for each token in a fill-in-the-blanks prompt for a $16$-way-$5$-shot setting using LLaMA-3.1$_\textnormal{\small{8B-Instruct}}$ and $3000$ RelSpec features per class. TokenScore highlights those tokens which were most relevant to the probes' final prediction of the relation type, which in this case matches the gold label (competition class). \textit{Note: Negative TokenScores are hidden and for attention weights we exclude the value for the first token from color scaling, as it is consistently larger, reducing visibility of other token values.}}
  \label{fig:appendix-fill-in-blanks-examples}
\end{figure}

\newpage
\section{Layerwise TokenScore and Attention Weights}
\label{fig:appendix-layerwise}
\begin{figure}[H]
  \centering

  \begin{subfigure}{\textwidth}
    \centering
    \includegraphics[height=0.55\linewidth]{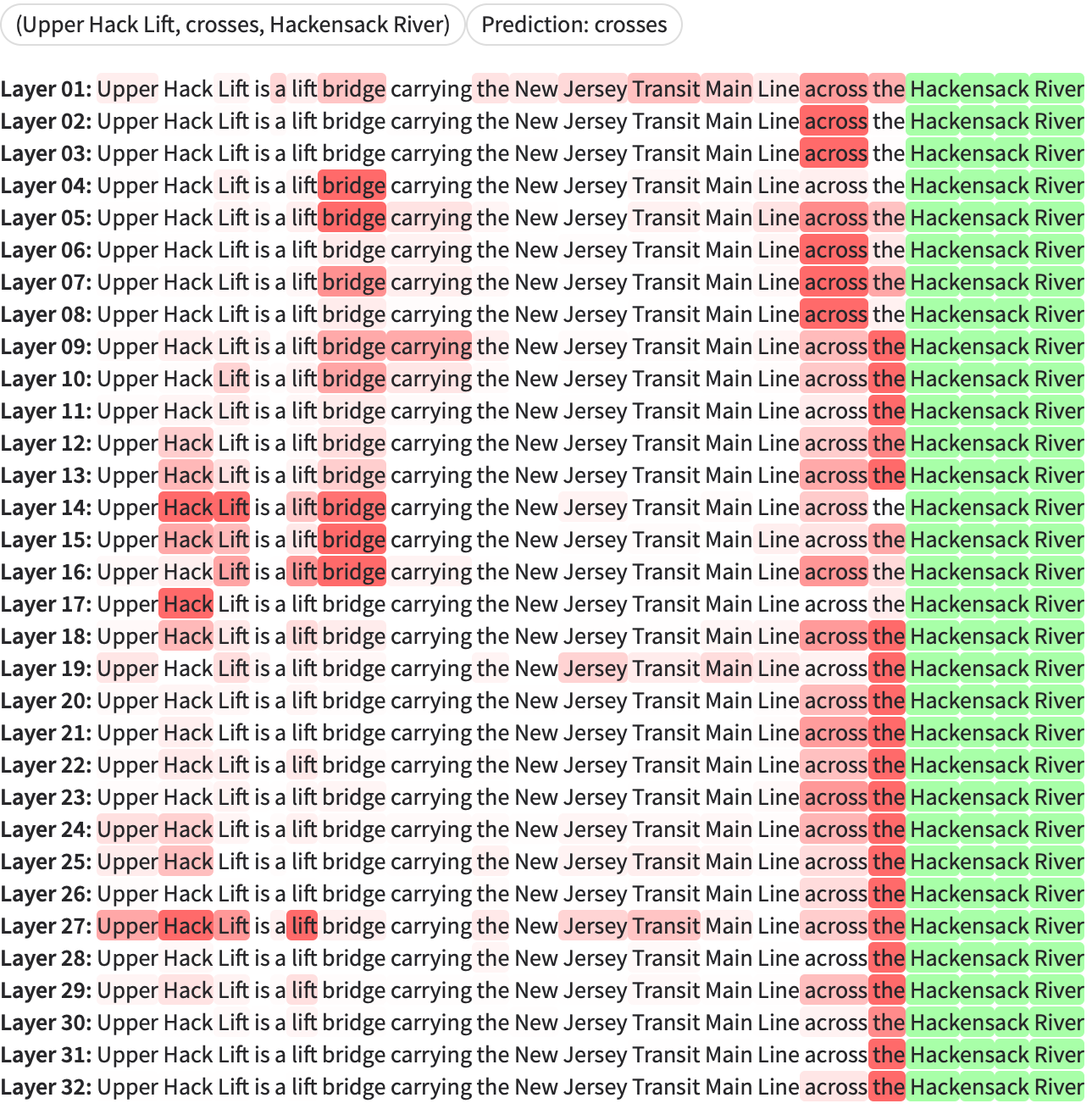}
    \caption{TokenScore.}
  \end{subfigure}
  \begin{subfigure}{\textwidth}
    \centering
    \includegraphics[height=0.55\linewidth]{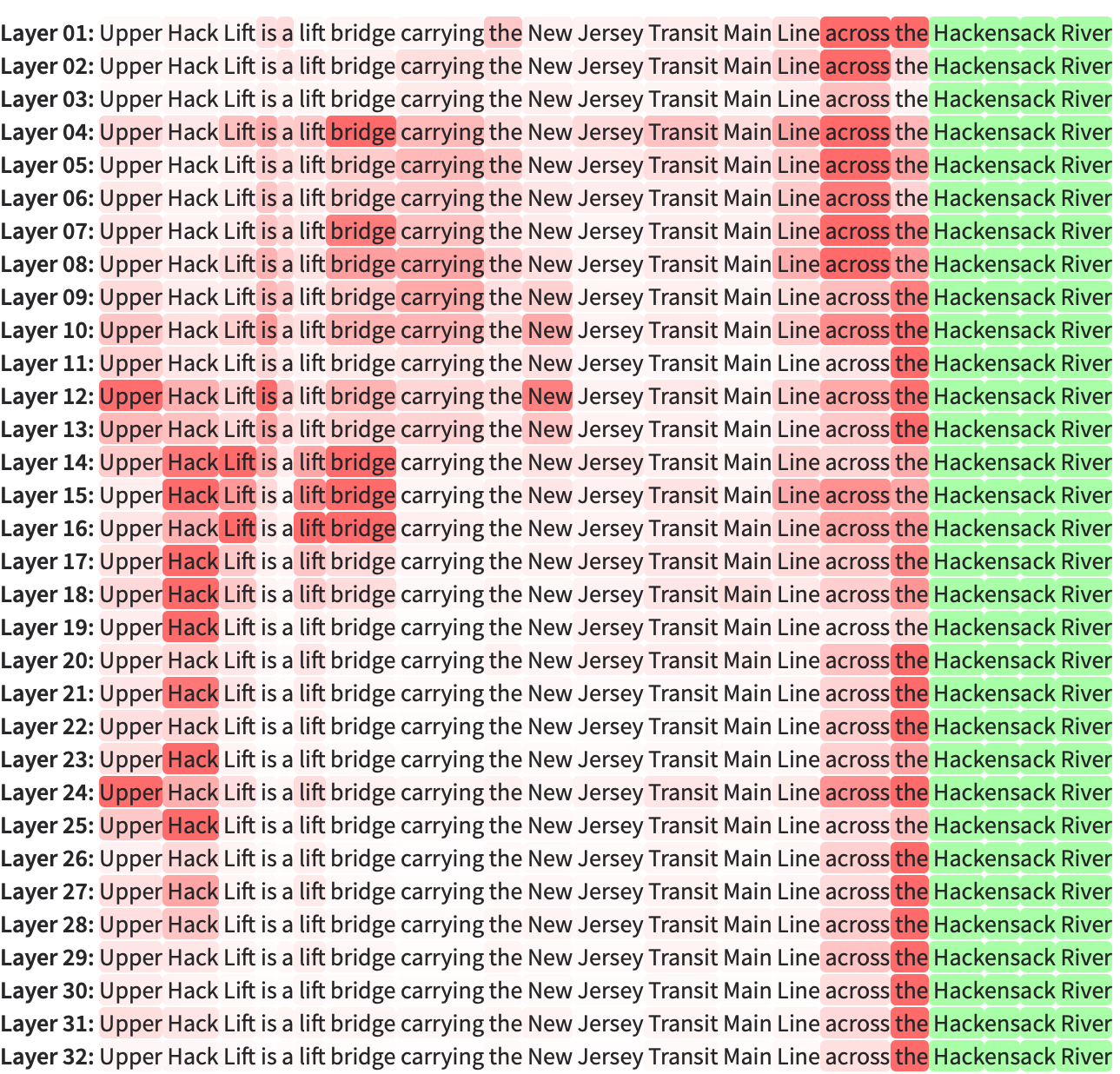}
    \caption{Attention weights.}
  \end{subfigure}\

  \caption{Shown are both the aggregated TokenScore, as well as the average attention weights \textit{per layer} for a $16$-way-$5$-shot query using LLaMA-3.1$_\textnormal{\small{8B-Instruct}}$ and $3000$ RelSpec features per class. TokenScore highlights those tokens which were most relevant to the probes' final prediction of the relation type, which in this case matches the gold label (competition class). \textit{Note: Negative TokenScores are hidden and for attention weights we exclude the value for the first token from color scaling, as it is consistently larger, reducing visibility of other token values. Shown are not the full prompts used, but only the context immediately before knowledge recall.}}
  \label{fig:appendix-layerwise-comparison}
\end{figure}

\begin{figure*}[htbp]
  \centering

  \begin{subfigure}{\textwidth}
    \centering
    \includegraphics[height=0.6\linewidth]{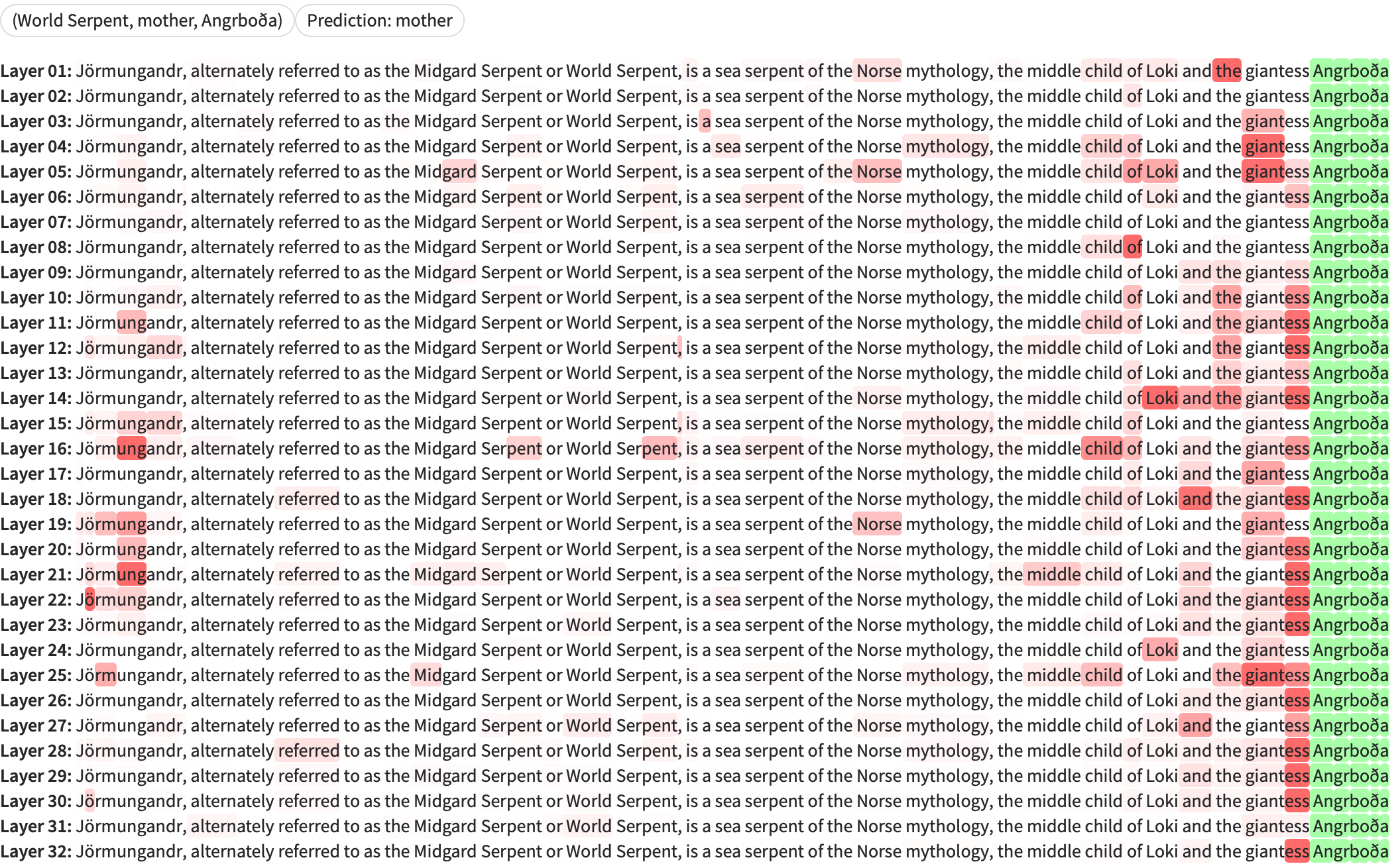}
    \caption{An example showing misalignment of subject entities according to annotation and according to attention head features: While the annotation calls for "World Serpent" as subject entity, TokenScores suggest that "Jörmungandr" is the span which the model attends to. The text itself lists "World Serpent" as an alias for "Jörmungandr", arguably making both interpretations correct. Interestingly, the focus does not seem to be on the last token of the subject span in this case. These inconsistencies are likely one reason why source-token dependent features, such as $\Delta_{\text{att},e_1}$ and $\Delta_{\text{att},e_1,h}$ do not perform well for classification.}
  \end{subfigure}
  \begin{subfigure}{\textwidth}
    \centering
    \includegraphics[height=0.6\linewidth]{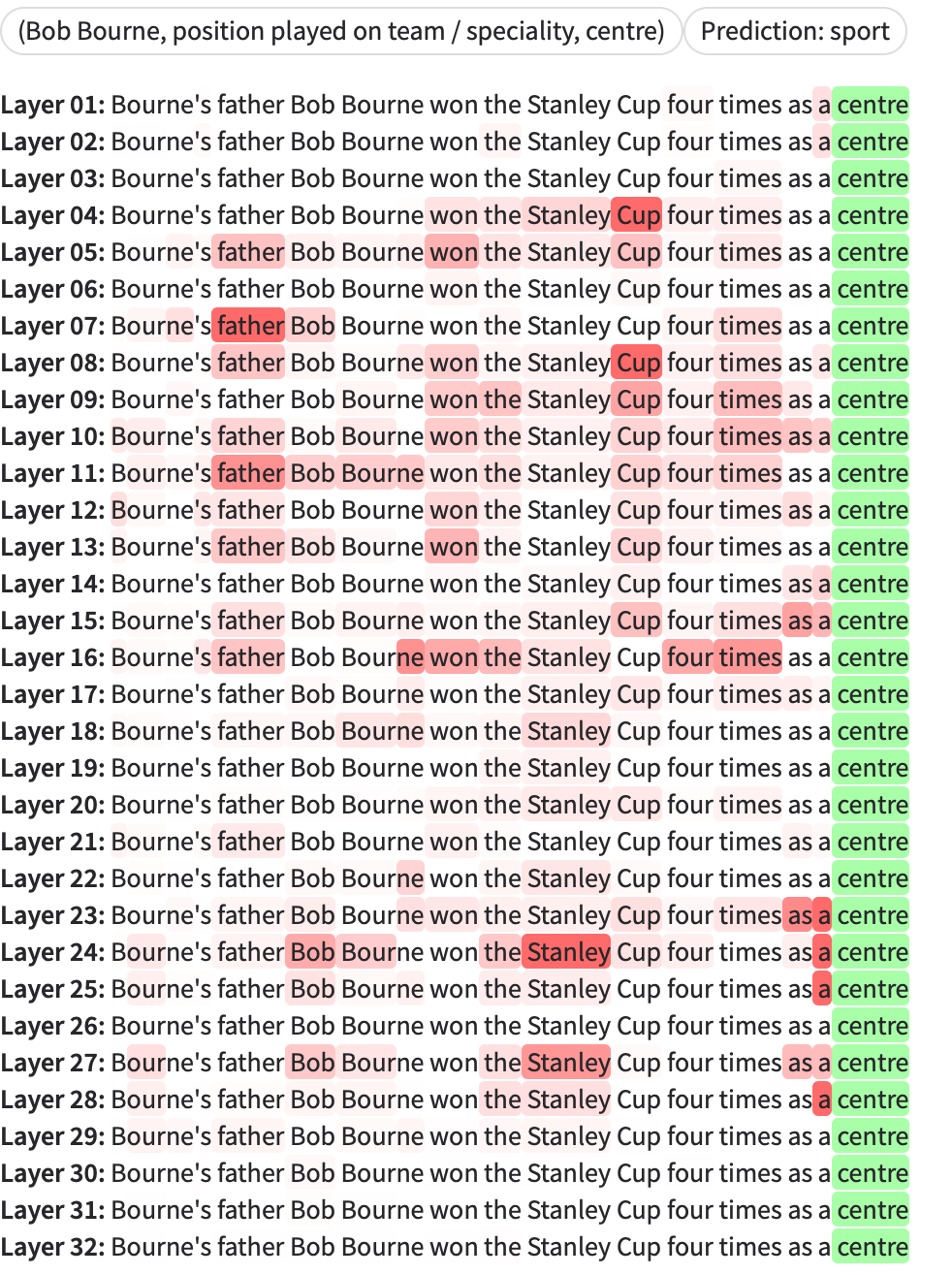}
    \caption{An example showing misalignment of subject entities according to annotation and according to attention head features: While the annotation calls for "Bob Bourne" as subject entity, TokenScores suggest that "Bob Bourne's father" is the span which the model attends to. Again both interpretations are technically correct. Furthermore, this example is a case of misclassification.}
  \end{subfigure}\

  \caption{Examples of per-layer TokenScores. All examples were created with fill-in-the-blanks prompt for a $16$-way-$5$-shot setting using LLaMA-3.1$_\textnormal{\small{8B-Instruct}}$. \textit{Note: Shown are not the full prompts used, but only the context immediately before knowledge recall.}}
  \label{fig:appendix-layerwise-observations}
\end{figure*}

Shown in Figure \ref{fig:appendix-layerwise-comparison} and Figure \ref{fig:appendix-layerwise-observations}.

\newpage

\section{Comparison of TokenScore and Gradient-based Analysis following \citet{geva-etal-2023-dissecting}}

\begin{figure*}[!ht]
    \centering
    \includegraphics[width=1\linewidth]{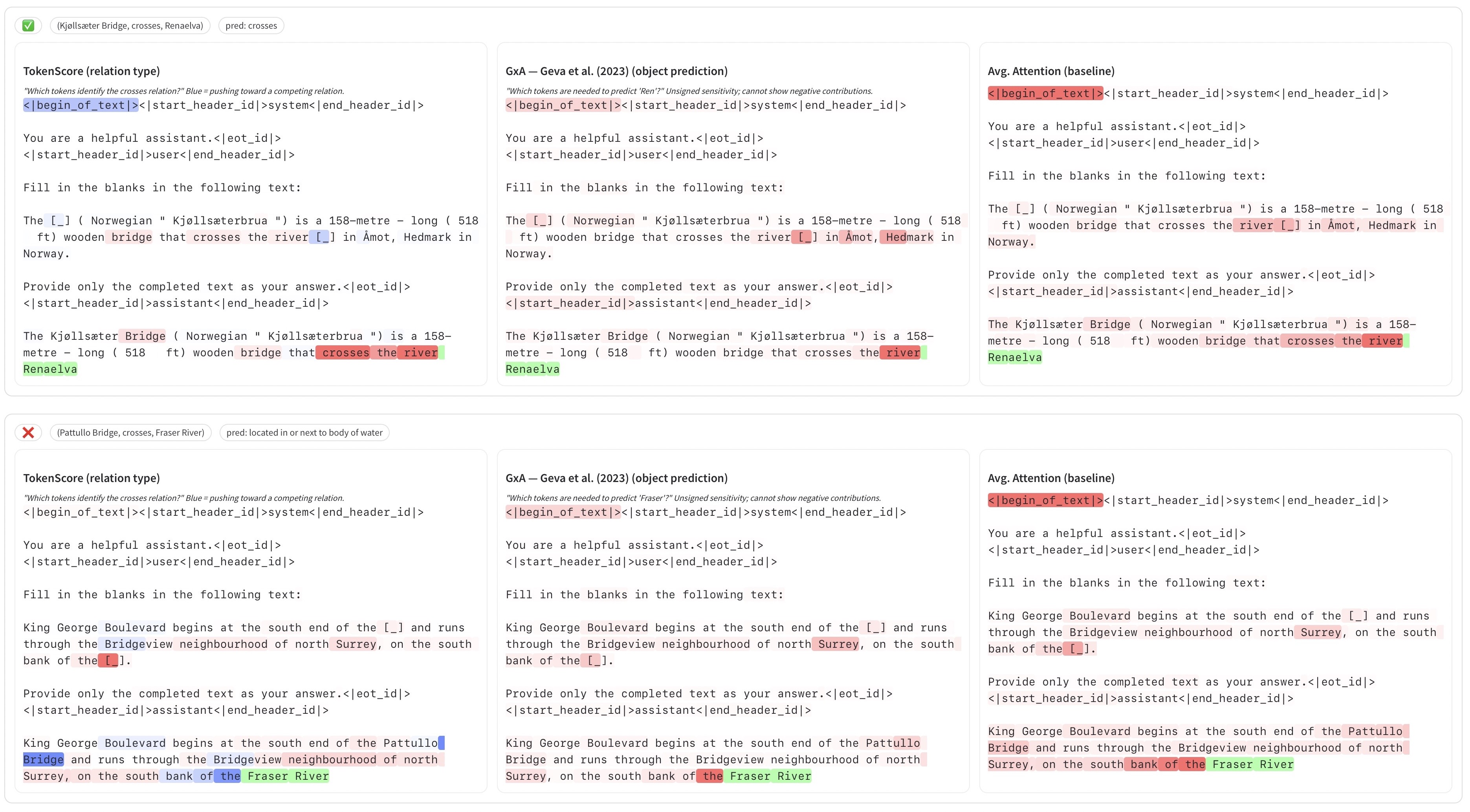}
    \caption{Two examples of probe predictions, one of correct, the other of an incorrect prediction for the relation type "crosses". Shown are TokenScore, GxA \cite{geva-etal-2023-dissecting}, and avg. attention weights. Produced in a $16$-way-$5$-shot setting using LLaMA-3.1$_\textnormal{\small{1B-Instruct}}$.}
    \label{fig:token_attribution}
\end{figure*}

\end{document}